\documentclass{article}

\usepackage{PRIMEarxiv}

\usepackage[utf8]{inputenc} 
\usepackage[T1]{fontenc}    
\usepackage{hyperref}       
\usepackage{url}            
\usepackage{booktabs}       
\usepackage{amsfonts}       
\usepackage{nicefrac}       
\usepackage{microtype}      
\usepackage{lipsum}
\usepackage{fancyhdr}       
\usepackage{graphicx}       
\graphicspath{{media/}}     
\usepackage{subfigure}
\usepackage{color}
\usepackage{caption}
\title{DynaMimicGen: A Data Generation Framework for Robot Learning of Dynamic Tasks
}

\author{
  Vincenzo Pomponi, Stefano Baraldo, Oliver Avram, Anna Valente \\
  Institute of Systems and Technologies for Sustainable Production (ISTePS) \\
  Department of Innovative Technologies (DTI) \\
  University of Applied Science and Arts of Southern Switzerland (SUPSI) \\
  Lugano, Switzerland \\
  \texttt{\{name.surname\}@supsi.ch} \\
  \And
  Loris Roveda \\
  Istituto Dalle Molle di studi sull’intelligenza artificiale (IDSIA) \\
  Department of Innovative Technologies (DTI) \\
  University of Applied Science and Arts of Southern Switzerland (SUPSI) \\
  Lugano, Switzerland\\
  \texttt{\{name.surname\}@supsi.ch} \\
  Department of Mechanical Engineering \\
  Politecnico di Milano (PoliMi) \\
  Milan, Italy\\
  \texttt{\{name.surname\}@mail.polimi.it}
   \And
  Paolo Franceschi \\
  Istituto Dalle Molle di studi sull’intelligenza artificiale (IDSIA) \\
  Department of Innovative Technologies (DTI) \\
  University of Applied Science and Arts of Southern Switzerland (SUPSI) \\
  Lugano, Switzerland\\
  \texttt{\{name.surname\}@supsi.ch} \\
   \And
  Luca Maria Gambardella \\
  Istituto Dalle Molle di studi sull’intelligenza artificiale (IDSIA) \\
  Faculty of Informatics \\
  Università della Svizzera Italiana (USI) \\
  Lugano, Switzerland \\
  \texttt{\{name.surname\}@usi.ch} \\
}

\begin{document}
\maketitle

\begin{abstract}
Learning robust manipulation policies typically requires large and diverse datasets, the collection of which is time-consuming, labor-intensive, and often impractical for dynamic environments.
In this work, we introduce \textit{DynaMimicGen} (D-MG), a scalable dataset generation framework that enables policy training from minimal human supervision while uniquely supporting dynamic task settings.
Given only a few human demonstrations, D-MG first segments the demonstrations into meaningful sub-tasks, then leverages Dynamic Movement Primitives (DMPs) to adapt and generalize the demonstrated behaviors to novel and dynamically changing environments.
Improving prior methods that rely on static assumptions or simplistic trajectory interpolation, D-MG produces smooth, realistic, and task-consistent Cartesian trajectories that adapt in real time to changes in object poses, robot states, or scene geometry during task execution.
Our method supports different scenarios — including scene layouts, object instances, and robot configurations — making it suitable for both static and highly dynamic manipulation tasks.
We show that robot agents trained via imitation learning on D-MG-generated data achieve strong performance across long-horizon and contact-rich benchmarks, including tasks like cube stacking and placing mugs in drawers, even under unpredictable environment changes.
By eliminating the need for extensive human demonstrations and enabling generalization in dynamic settings, D-MG offers a powerful and efficient alternative to manual data collection, paving the way toward scalable, autonomous robot learning.
GitHub project at \href{https://github.com/automation-robotics-machines/DynaMimicGen}{D-MG}
\end{abstract}


\section{Introduction}
Learning from Demonstration (LfD) \cite{Billard2008} is a methodology used to transfer new skills to machines by relying on demonstrations provided by a user.
This approach aims to make robot programming more intuitive and accessible \cite{franceschi2023modeling,VALENTE202221,10731373,10363056,franceschi2025human}, especially for novice users, by enabling robots to learn directly from human demonstrations without requiring extensive programming expertise \cite{Calinon2018}.
However, traditional Imitation Learning (IL) \cite{kroemer2021review} methods typically require a large number of human-provided demonstrations to generalize across different scene configurations, object instances, and robotic platforms \cite{12345678902345}.
In addition, collecting such datasets is time-consuming, labor-intensive, and often impractical for real-world applications \cite{osa2018algorithmic}.
To overcome this limitation, we propose \textbf{D-MG}, a novel framework that leverages Dynamic Movement Primitives (DMPs) to generate large-scale and diverse datasets in dynamic environments with minimal human supervision, often requiring a single demonstration for most of the tasks.

\textbf{D-MG} introduces a novel framework for robot dataset generation that transforms a minimal set of human demonstrations—potentially as few as a single example—into diverse and realistic robot trajectories expressed in absolute Cartesian end-effector space.
By encoding demonstrated behaviors using Dynamic Movement Primitives (DMPs)  \cite{dmps_01, dmps_02, dmps_03}, D-MG enables smooth trajectory modulation that preserves the essential dynamics of each subtask while adapting to new, possibly dynamic, scene configurations.
This capability allows the synthesis of large-scale datasets that reflect variations in scene geometry, object placements, and robot embodiments without requiring extensive manual supervision.

Prior works typically assume quasi-static environments, where object poses remain fixed or can only be observed at the beginning of each subtask. As a result, they struggle to generalize to dynamic task settings.
In this work, we address a broader and more practical question: \textbf{How can we synthesize large-scale robotic datasets that allow deep learning policies to generalize effectively during deployment?}

Unlike previous methods, D-MG does not assume static scenes.
It continuously senses object poses during trajectory synthesis, enabling online adaptation to changing goal configurations.
Rather than relying on variability present in the human demonstrations, D-MG introduces structured environmental perturbations to increase dataset diversity and strengthening the robustness of learned policies.

Through this capability, D-MG extends synthetic data generation to complex, unstructured settings and strengthens the link between human demonstrations and effective downstream policy learning.

The key contributions of this paper are:
\begin{itemize}
    \item \textbf{scalable data generation from minimal input:} we introduce \textbf{D-MG}, a novel framework for generating large-scale robot manipulation datasets from minimal human input—often as little as a single demonstration. By leveraging Dynamic Movement Primitives (DMPs), D-MG generalizes demonstrated behaviors to novel scenes, drastically reducing the need for extensive human data collection;

    \item \textbf{data generation in dynamic task settings:} In contrast to prior methods, D-MG enables data generation in dynamic and evolving environments. Leveraging the adaptability of Dynamic Movement Primitives (DMPs), our framework can adjust motion plans in real time to accommodate changes in object position or orientation during task execution. Importantly, this robustness is achieved collecting human demonstrations in static scenes;

    \item \textbf{valuable data generation for imitation learning agents:} we demonstrate that DynaMimicGen can generate high-quality data to train proficient agents via imitation learning across diverse scene configurations, all of which are unseen in the original demos.
\end{itemize}  

By addressing these challenges, D-MG represents a step toward more scalable and efficient data-driven robot learning, enabling robots to autonomously generalize task execution from minimal human input.

\section{Related Works}
Dynamic Movement Primitives (DMPs) \cite{dmps_01, dmps_02, dmps_03} encode demonstrations as nonlinear second-order dynamical systems with a learnable forcing term, ensuring convergence to a goal while allowing flexible trajectory shapes.
By adjusting the goal parameter, the same motion pattern can generalize to new target positions \cite{5152385}.
Extensions to the DMP framework further enhance its expressiveness.
\cite{5152385} leverage DMPs to construct libraries of reusable manipulation skills (e.g., grasp, place, release), enabling complex task execution by sequencing primitives.
\cite{5152423} incorporate repulsive potential fields to facilitate obstacle-aware motion, while \cite{6748918} propose Coupling Movement Primitives to allow coordination among multiple DMPs (e.g., for bimanual tasks) and online modulation of temporal and amplitude parameters.

Despite their adaptability and long-standing use in imitation learning, no prior work has explored synthesizing large-scale robotic manipulation datasets using DMPs.

Recent advances in human-robot collaboration \cite{roveda2021human} and imitation learning emphasize the importance of scalable synthetic data generation \cite{levine2018learning, pinto2016supersizing, open_x_embodiment_rt_x_2023, brohan2023rt} to train robust manipulation policies from minimal human supervision.
Several works explore data-driven synthesis of demonstrations to improve generalization.
MimicGen \cite{mandlekar2023mimicgen} and DexMimicGen \cite{jiang2024dexmimicgen} segment human trajectories into object-centric subtasks and replay them after rigid SE(3) transformations, but they assume static scenes and use linear interpolation, which can lead to collisions under changing conditions.
Similarly, DemoGen \cite{xue2025demogen} and DreMa \cite{barcellona2024dream} rearrange static scenes to augment data but remain limited to quasi-static tasks, while IntervenGen \cite{hoque2024intervengen} introduces closed-loop corrections without fully addressing dynamic scene variation.

Overall, current demonstration-based augmentation approaches rely on rigid, scene-static assumptions that limit their adaptability to dynamic environments.

At the same time, large-scale imitation learning (IL) efforts demonstrate that data diversity and volume are key to generalization.
Early self-supervised grasping studies collected tens of thousands of trials via trial-and-error or teleoperation \cite{pinto2016supersizing, Calinon2018}, showing that performance scales with data size rather than complex reward shaping.
Subsequent initiatives such as RoboNet \cite{dasari2019robonet} aggregated over 160K trajectories from multiple robots and environments, enabling cross-domain transfer.
More recently, RT-1 \cite{brohan2022rt} and the Open X-Embodiment initiative \cite{open_x_embodiment_rt_x_2023} leveraged millions of real-world demonstrations across heterogeneous robots, highlighting that large, multi-robot datasets can produce general-purpose manipulation policies.
Collectively, these studies underscore a consistent trend: scalable imitation learning emerges from massive, heterogeneous data aggregation, where diversity acts as a substitute for handcrafted supervision.

Finally, D-MG addresses all the aforementioned limitations.
Starting from a single human demonstration, D-MG segments the task into subtasks and learns a dedicated DMP for each component, extracting the corresponding parameters. The framework then generates a potentially unbounded set of trajectories by generalizing the demonstrated motion to diverse object poses within the environment.
During execution, the DMP formulation enables continuous adaptation to dynamic changes in object position and orientation, ensuring consistent task completion.
This process yields large-scale trajectory variability from only one demonstration, enriching the dataset to train IL models.
By incorporating dynamic modulation, D-MG produces diverse trajectory shapes, improving model training performance.

\section{Problem setup}
\label{sec:problem}

\begin{figure}[h]
    \centering
    \includegraphics[width=1.0\linewidth]{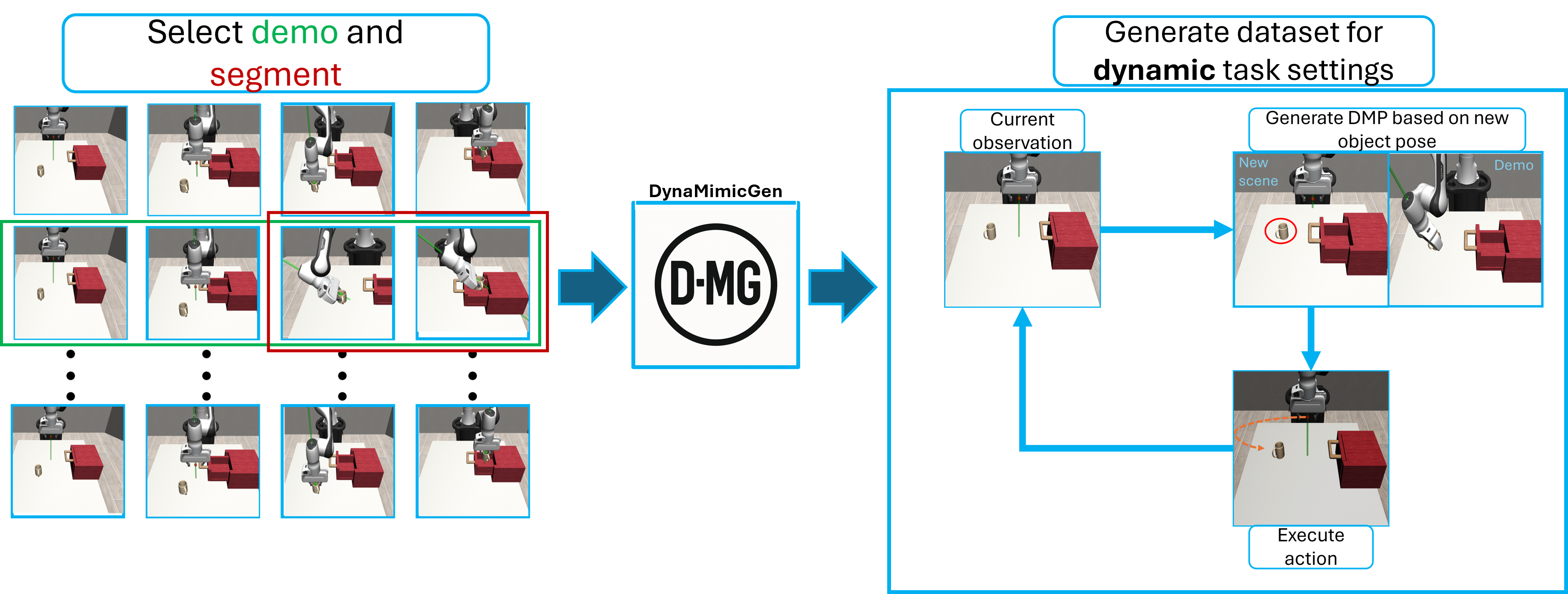}
    \caption{Overview of the DynaMimicGen system pipeline. \textbf{(Left)} DynaMimicGen begins by selecting the most relevant demonstration (highlighted in green) from the source dataset for the target task. It then identifies the appropriate reference segment (highlighted in red), corresponding to an object-centric subtask, to train a Dynamic Movement Primitive (DMP). \textbf{(Right)} To generate a new demonstration in a novel scene, DynaMimicGen (1) monitors the current state of the environment, (2) transforms the selected segment to generate the DMP goal for the current state configuration, and (3) executes the resulting trajectory. This process is performed with real-time monitoring of the environment to adapt the trajectory in response to dynamic changes.}
    \label{fig:D-MG framework}
\end{figure}

We formulate each robot manipulation task as a Markov Decision Process (MDP) \cite{rl_intro_barto} and aim to generate a dataset suitable for learning a robot manipulation policy $\pi$, which maps the state space $\mathcal{S}$ to the action space $\mathcal{A}$.
The imitation dataset consists of $N$ demonstrations, with $N \geq 1$, defined as $\mathcal{D} = \{(s_0^i, a_0^i, s_1^i, a_1^i, \dots, s_{H_i}^i)\}_{i=1}^N$, where each initial state $s_0^i$ is sampled from the initial state distribution $D$, \textit{i.e.}, $s_0^i \sim D(\cdot)$.
In this work we use Diffusion Policy \cite{chi2024diffusionpolicy} and Behavior Cloning with an RNN-based policy~\cite{mandlekar2021matters} to train the policies.
We evaluate our method by comparing its performance against MimicGen (MG) \cite{mandlekar2023mimicgen}.
For fairness, we reprocess the MG-generated delta trajectories into absolute coordinates, as D-MG operates on absolute trajectories and DMPs require full trajectory information for training.
Accordingly, we do not directly use the performance metrics reported in the original MG paper, but instead recompute them under this consistent evaluation setting (more details in Appendix \ref{app:faq}).

\subsection{Problem statement and assumptions}
Our objective is to generate a large dataset $\mathcal{D}_{gen}$ from a source dataset $\mathcal{D}_{src}$, composed of one or a two human demonstrations for a given task $\mathcal{M}$.
This dataset can be used to train policies for the original task or its variants, including changes in initial conditions.
Unlike prior work \cite{mandlekar2023mimicgen}, our method also supports dynamic environments by adapting trajectories in real time when object displacements are detected.

The data generation process proceeds as follows: (1) sample a new initial state from the target task distribution; (2) select an appropriate demonstration from $\mathcal{D}_{src}$ based on the new state; (3) segment the demonstration into object-centric motion primitives; (4) train a Dynamic Movement Primitive (DMP) for each segment; (5) adapt the DMPs to the current robot and object configuration; (6) execute the adapted actions while monitoring for scene changes; and (7) if successful, add the resulting trajectory to the generated dataset $\mathcal{D}_{gen}$.

D-MG makes the following assumptions:

\textbf{Assumption 1: absolute end-effector pose-action space.} 
The action space $\mathcal{A}$ is defined as a combination of absolute end-effector pose commands and a binary gripper open/close command.
The use of absolute pose commands is essential for effectively training Dynamic Movement Primitives (DMPs), as it allows the learned primitives to generalize and adapt consistently to new scenes sampled from the target task distribution.

\textbf{Assumption 2: tasks consist of a known sequence of object-centric subtasks}, \textit{i.e.}, the task planning is known.
Let $O = \{o_1, \cdots, o_K\}$ be the set of objects in a task $\mathcal{T}$.
As in \cite{di2022learning}, we assume that tasks consist of a sequence of object-centric subtasks $\big(S_1(o_{S_1}), S_2(o_{S_2}), \cdots, S_M(o_{S_M})\big)$, where the manipulation in each sub-task $S_i(o_{S_i})$ is relative to a single object's coordinate frame $(o_{S_i} \in O)$.
We assume this sequence is known.

\textbf{Assumption 3: object poses can be observed at each time-step during motion generation.}
We assume that we can observe the pose of the relevant object $o_{S_i}$ at any time step $t$ of each sub-task $S_i(o_{S_i})$ during data collection.
This assumption enables D-MG to generate consistent and adaptable trajectories taking into account dynamic changes in the environment (\textit{i.e.}, object poses).

\section{Method}
\label{sec:method}
In this section, we describe how our method generates new robot trajectories using a source set of $N \geq 1$ human demonstrations for a given task.
As shown in Figure \ref{fig:D-MG framework}, D-MG begins by parsing each source demonstration into segments, where each segment corresponds to an object-centric sub-task.
To generate a successful trajectory in a new scene, D-MG first selects the most appropriate demonstration from the source dataset and trains a corresponding Dynamic Movement Primitive (DMP) for each segment.
The robot executes this adapted sequence in the new scene using an end-effector controller, resulting in a coherent and scene-consistent trajectory.

\subsection{Parsing human demonstrations into object-centric segments}
\label{subsec:parsing}
We assume each manipulation task consists of a known, ordered sequence of object-centric subtasks, inspired by prior work on object-centric learning from demonstrations \cite{di2022learning, mandlekar2023mimicgen}.
Let $O = {o_1, \dots, o_K}$ denote the set of task-relevant objects.
Each task is modeled as a sequence of subtasks $\big(S_1(o_{S_1}), S_2(o_{S_2}), \dots, S_M(o_{S_M})\big)$, where each subtask $S_i$ is defined relative to the coordinate frame of a specific object $(o_{S_i} \in O)$.

For example, placing a mug inside a drawer can be decomposed into: (i) opening the drawer (motion relative to the drawer frame), (ii) grasping the mug (relative to the mug), (iii) placing the mug inside the drawer (relative to the drawer), and (iv) closing the drawer (relative to the drawer).
Sequencing these object-centric motions enables structured execution of complex tasks and serves as the basis for our data-generation pipeline.

To impose this segmentation, each demonstration trajectory $\tau$ is decomposed into contiguous segments $\tau = (\tau_1, \tau_2, \dots, \tau_M)$, where each $\tau_i$ corresponds to a subtask.
We leverage subtask-completion metrics that automatically detect transitions between subtasks using robot and object state signals—e.g., detecting mug grasp, mug placement, drawer opening, and drawer closure events.
Such signals are readily available in simulation, where task success is routinely monitored.

In real-world settings where privileged state information may be unavailable, subtask boundaries can instead be manually annotated during or after demonstration collection, which remains feasible given the typically modest number of expert demonstrations.
In our experiments, we rely on automated metrics that were either directly provided by the environment or straightforward to design for each task.

\subsection{Generating a DMP for Each Sub-Task in a New Scene}
\label{subsec:DMP_gen}
In this section, we provide an overview of Dynamic Movement Primitives (DMPs), a widely used method for representing and generalizing robot trajectories.

DMPs model motion as a second-order dynamical system, where a nonlinear forcing term allows the system to reproduce complex trajectories while maintaining stability.
This formulation enables the robot to adapt a demonstrated movement to new start and goal configurations.

The DMPs used in this work rely on a second-order system described by \cite{dmps_01, dmps_02, dmps_03}:
\begin{equation}
    \label{eq: dmp_equation}
    \ddot{y}(t)=\alpha_y(\beta_y(g-y(t))-\dot{y}(t))+f(x),
\end{equation}
where $y(t)$ is the system state, $\dot{y}(t)$ and $\ddot{y}(t)$ its first and second time derivatives, $\alpha_y$ and $\beta_y$ positive constants, $g$ the goal position, and $f(x)$ is a forcing term.
The forcing term is computed as:
\begin{equation}
    f(x) = \frac{\sum_{i=1}^{K} \Psi_i(x) \omega_i }{\sum_{i=1}^{K}  \Psi_i(x)} x(g-y_0),
\end{equation}
where $y_0$ is the initial position of the system, $\omega_i$ are weights, $\Psi_i(x)$ are fixed $K$ Gaussian basis functions:
\begin{equation}\label{equ:psi}
    \Psi_i(x) = exp(-h_i(x-c_i)^2),
\end{equation}
and $x$ is the so called \textit{canonical system}, defined as:
\begin{equation}\label{equ:cansys}
    \dot{x} = \alpha_x x.
\end{equation}

The weights $\omega_i$ must be learned from the demonstration, to determine its representation as a DMP.
The standard method used to learn them is \textit{Locally Weighted Regression} (LWR), briefly described in the following.
Given a demonstration trajectory $\tau_{demo} = \{y_d(t_0), y_d(t_1), \dots y_d(t_f)\}$, equation \ref{eq: dmp_equation} is inverted and $f_d$ computed as 
\begin{equation}\label{equ:f_d}
    f_d(t) = \ddot{y_d}(t) - \alpha_y(\beta_y(g-y_d(t))-\dot{y_d}(t)).
\end{equation}
A function approximation problem is formulated to find the $\omega_i$ parameters that make $f_d$ as close as possible to $f$.
For each kernel function $\Psi_i$, LWR looks for the corresponding $\omega_i$ that minimizes the locally weighted quadratic error through the following cost function
\begin{equation}\label{equ:Ji}
    J_i = \sum_{t=1}^P \Psi_i(t)(f_d(t)-\omega_i(x(g-y_0))).
\end{equation}

Once the kernel parameters $\omega_i$ for each basis function $\Psi_i$ are learned, the system can reproduce the demonstrated motion while adapting to new start and goal conditions.
The resulting DMP generates smooth trajectories that retain the essential characteristics of the demonstration, ensuring both stability and flexibility.
This capability makes DMPs well-suited for tasks that require generalization and online adaptation, such as the dynamic manipulation scenarios considered in this work.

\subsection{Adapting the source sub-task segment.}
\label{subsec:seg_adaptation}
Each sub-task segment $\tau_i$ is interpreted as a sequence of target poses for the end-effector controller (see Assumption 1, Section \ref{sec:problem}).
Let $T_B^A$ denote the homogeneous transformation matrix representing the pose of frame $A$ with respect to frame $B$.
Then, a sub-task trajectory segment can be expressed as
\begin{equation}
    \tau_i = (T_W^1, \dots, T_W^{K}),
\end{equation}
where $T_W^k$ denotes the target homogeneous matrix at timestep $t_k$, $W$ is the world frame, and $K$ is the segment length.

Since each motion is assumed to originate from the robot's current configuration and terminate at a goal pose relative to the manipulated object, the DMP trajectory is generated based on the initial configuration of the robot and the object's pose in the new scene.
The new start pose of the DMP is set as the robot’s configuration in the new scene.
Determining the goal pose requires additional computation, which we illustrate in the following.

Let $\hat{T}_W^{obj}$ denote the homogeneous transformation matrix of the object relative to the world frame in the demonstration, and let $\hat{T}_W^{target}$ represent the end-effector pose exactly at the final time step of the sub-task.
Therefore, we compute:
\begin{equation}
    \hat{T}_{obj}^{target} = \left( \hat{T}_W^{obj} \right)^{-1} \hat{T}_W^{target}.
\end{equation}
which is the homogeneous transformation matrix of the end-effector with respect to the object reference frame.
Thus, in the new scene, the target end-effector pose results in 
\begin{equation}
    \hat{T}_W^{target} = \tilde{T}_W^{obj} \hat{T}_{obj}^{target},
\end{equation}
where $\tilde{T}_W^{obj}$ represents the homogeneous transformation matrix of the object relative to the world frame in the updated environment state.
From $\tilde{T}_W^{target}$, the corresponding updated target pose for the end-effector can then be extracted.

\subsection{Executing the updated segment.}
\label{subsec:seg_execution}
DynaMimicGen (D-MG) continuously monitors object positions during execution and dynamically adapts the end-effector trajectory using trained DMPs.
At each timestep, it samples the updated target pose, pairs it with the corresponding gripper action from the provided demonstration, and issues control commands.
This process repeats across all task segments, with only successful executions retained after post-execution validation.
The Data Generation Rate (DGR) is defined as the proportion of successful trials. Relying solely on object and controller frames, D-MG generalizes across tasks with varying initial states, object instances, and robots.
In experiments, we evaluate this flexibility by varying the initial distributions (D), for each representative task.

\section{Experiment Setup}
\label{sec:exp_setup}

\begin{figure}
    \centering
    \subfigure[\centering Stack]{\includegraphics[width=1.0\linewidth]{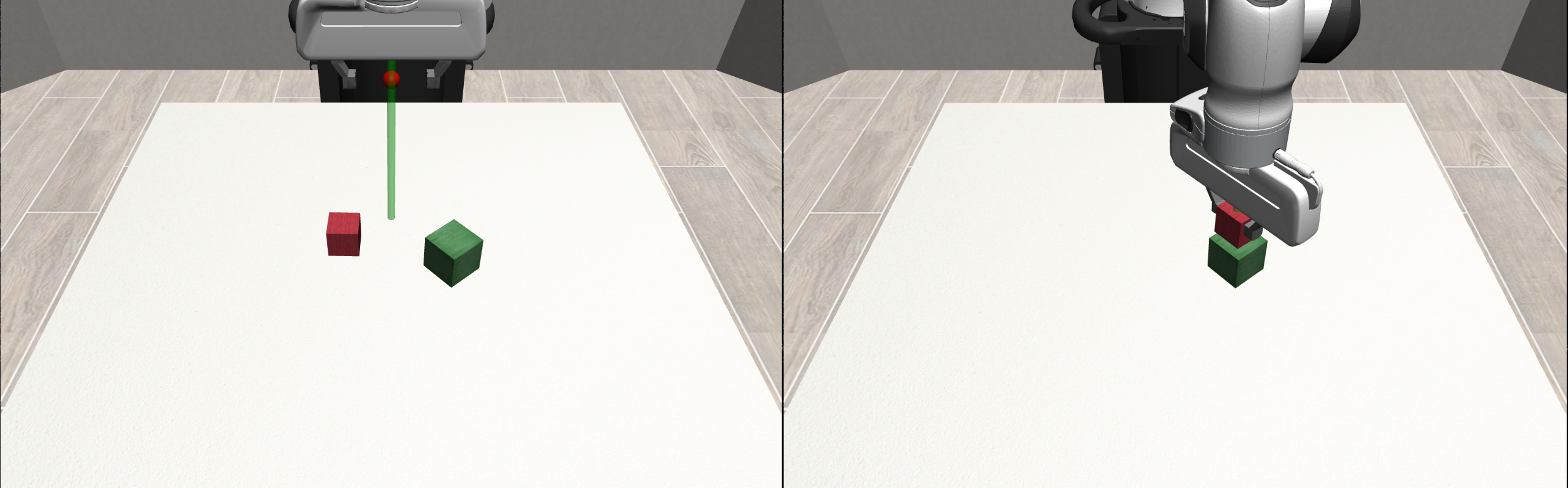}}\\
    \subfigure[\centering Square]{\includegraphics[width=1.0\linewidth]{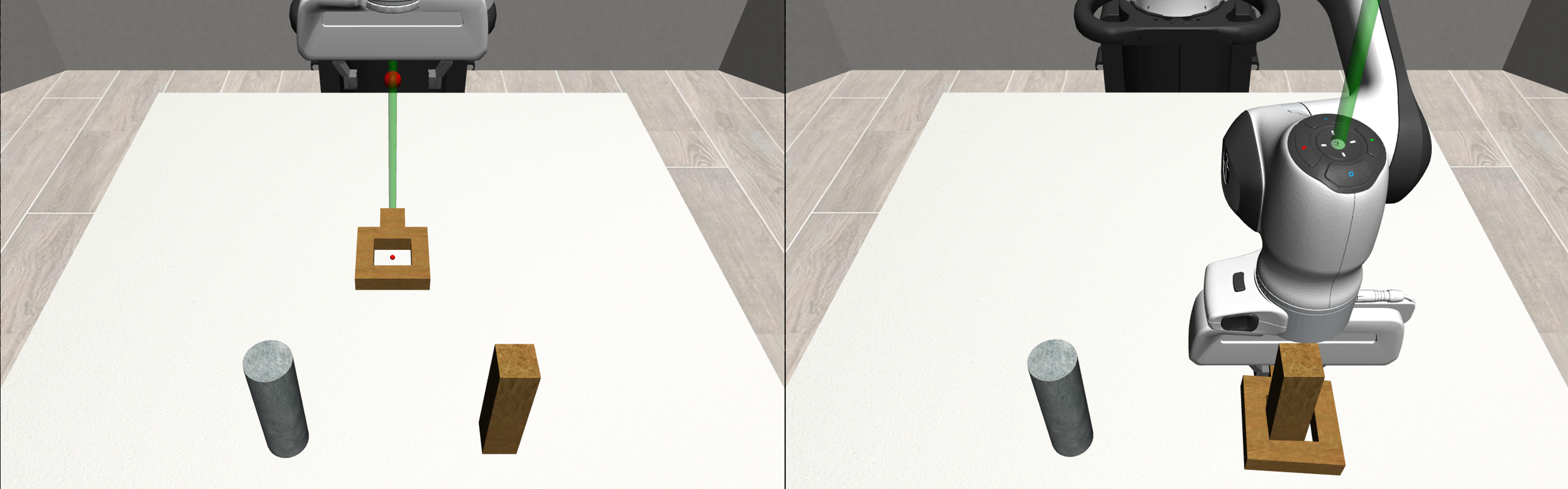}}\\
    \subfigure[\centering MugCleanup]{\includegraphics[width=1.0\linewidth]{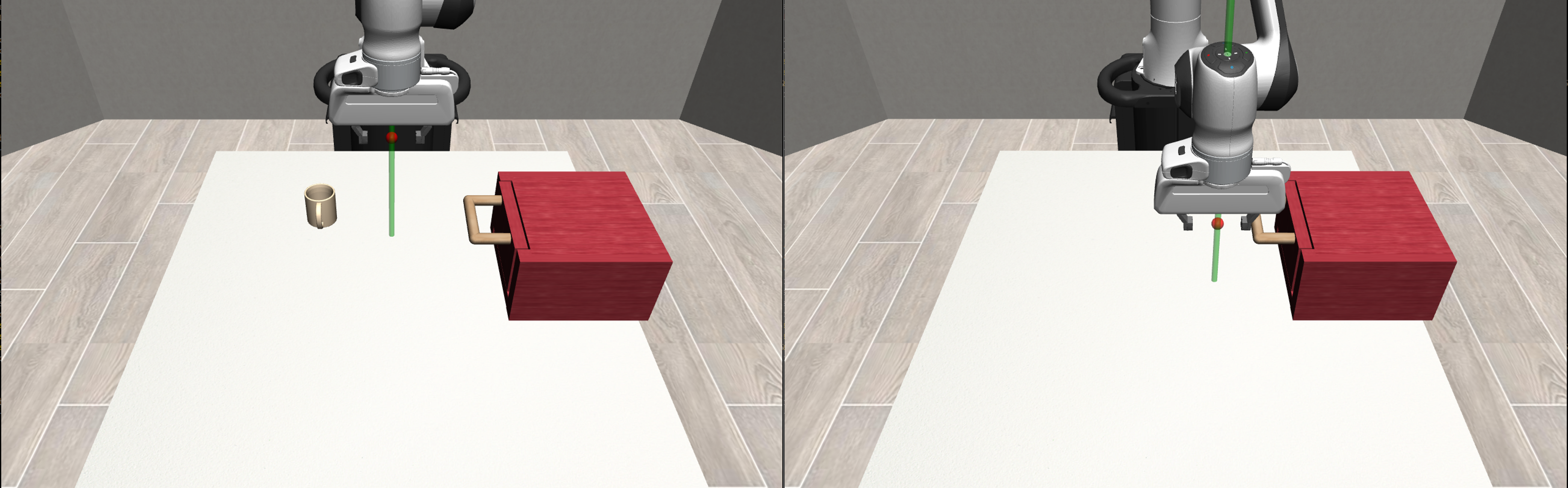}}
    \caption{\textbf{Tasks.} We show all of the simulation tasks in the figure above. They span different behaviors including pick-and-place, precise insertion and articulation, and include long-horizon tasks requiring chaining several behaviors together.}
    \label{fig:AllTasks}
\end{figure}

We evaluate D-MG across a diverse suite of manipulation tasks, each representing a distinct behavioral category to demonstrate its versatility in generating useful imitation-learning data.
Specifically, we consider: (i) Basic manipulation (Stack), (ii) Contact-rich interaction (Square), and (iii) Long-horizon sequential manipulation (MugCleanup).\\
\subsection{Tasks and Task Variants}
\label{subsec:tasks_and_variants}

The tasks are clearly shown in Figure \ref{fig:AllTasks}.
We group the tasks into categories as in Sec. \ref{sec:problem} and describe the goal, the variants, and the object-centric subtasks in each task.
As mentioned in Sec. \ref{sec:method}, the action space $\mathcal{A}$ is defined as a combination of absolute end-effector pose commands and a binary gripper open/close command (implemented with an Operational Space Controller \cite{1087068}).
Control happens at 20 hz.\\
\textbf{Basic}. A task involving pick-and-place motions, minimal contacts and simple spatial relationships.
\begin{itemize}
    \item \textbf{Stack} \cite{zhu2020robosuite} Stack a red block on a green one.
    Blocks are initialized in a small (0.16m x 0.16m) region ($D_0$) and a large (0.4m x 0.4m) region ($D_1$) with a random top-down rotation.
    There are 2 subtasks (grasp red block, place onto green).
    We also develop a version of this task in the real-world (Fig. \ref{fig:real_envs}).
\end{itemize}
\textbf{Contact-Rich}. A task involving contact-rich behaviors such as insertion or drawer articulation.
\begin{itemize}
    \item \textbf{Square}. Pick a square nut and place on a peg.
    (D0) Peg never moves, nut is placed in small (0.005m x 0.115m) region with a random top-down rotation.
    (D1) Peg and nut move in large regions, but peg rotation fixed. Peg is initialized in 0.4m x 0.4m box and nut is initialized in 0.23m x 0.51m box.
    (D2) Peg and nut move in larger regions (0.5m x 0.5m box of initialization for both) and peg rotation also varies.
    There are 2 subtasks (grasp nut, place onto peg).
\end{itemize}
\textbf{Long-Horizon}. A task requiring chaining multiple behaviors together.
\begin{itemize}
    \item \textbf{Mug Cleanup}.
    The MugCleanup task consists of four subtasks — opening the drawer, grasping the mug, placing it inside the drawer, and closing the drawer — along with additional task variants.
    ($D_0$) The drawer does not move and the mug moves in a 0.3m x 0.15m box with a random top-down rotation.
    ($D_1$) The mug moves in a 0.2m x 0.1m box with 60 degrees of topdown rotation variation and the mug is initialized in a 0.4m x 0.15m box with a random top-down rotation.
\end{itemize}

\begin{figure}[h]
    \centering
    \subfigure[\centering]{\includegraphics[width=8.0cm]{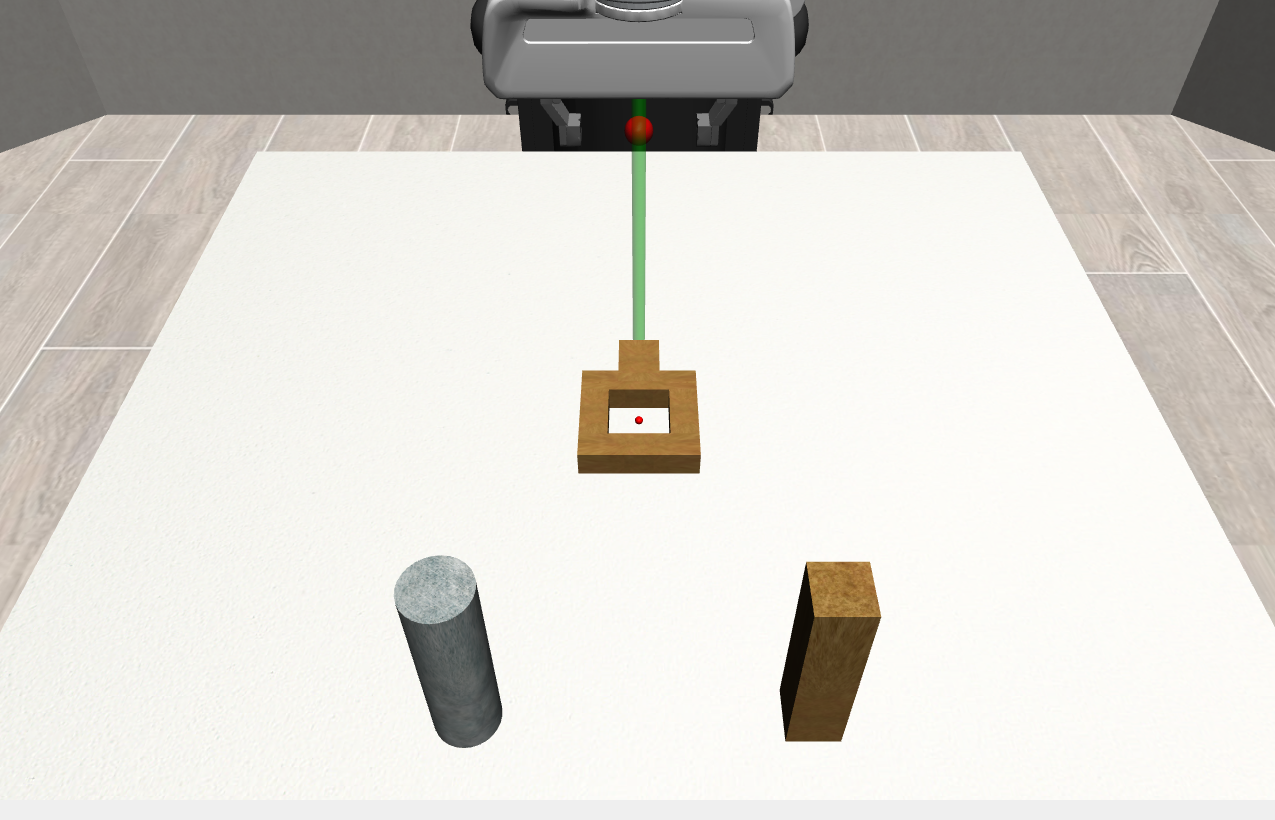}}
    \subfigure[\centering]{\includegraphics[width=8.0cm]{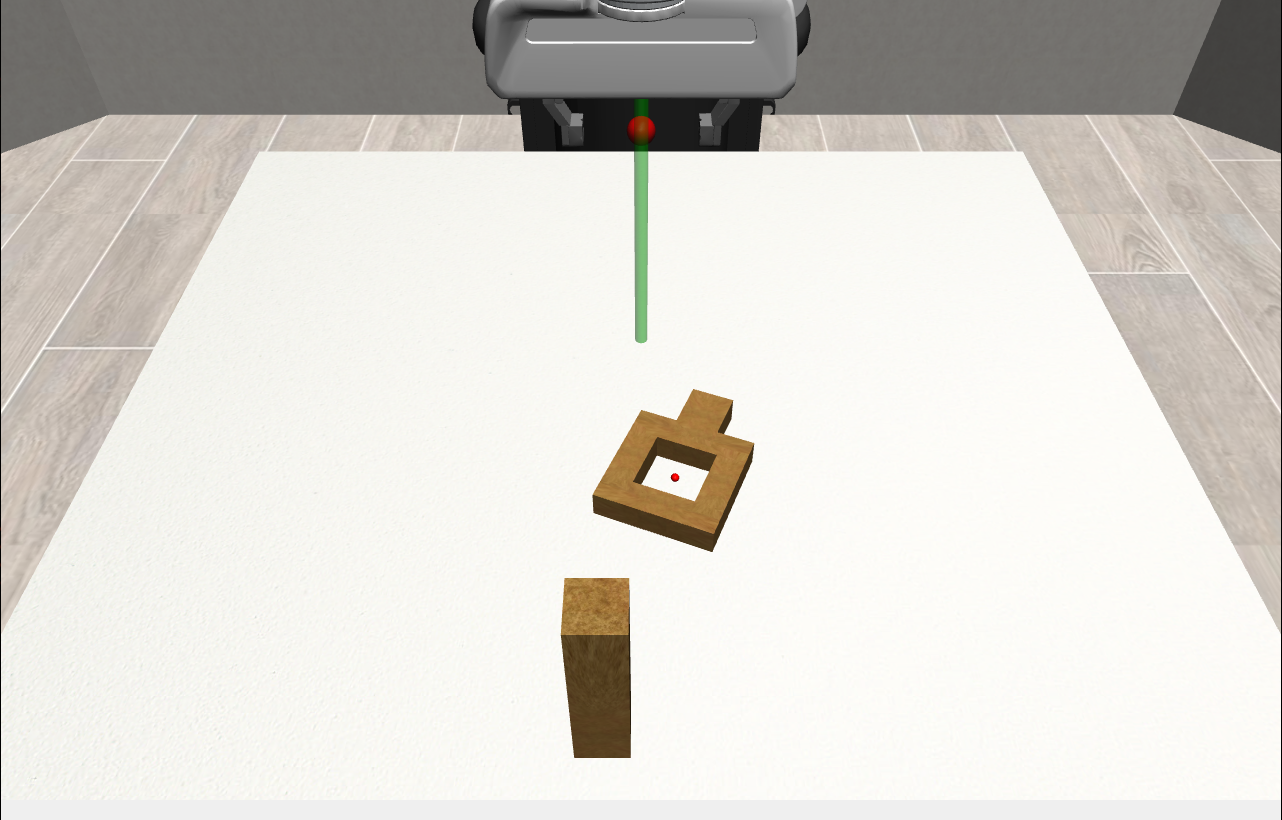}}\\
    \subfigure[\centering]{\includegraphics[width=8.0cm]{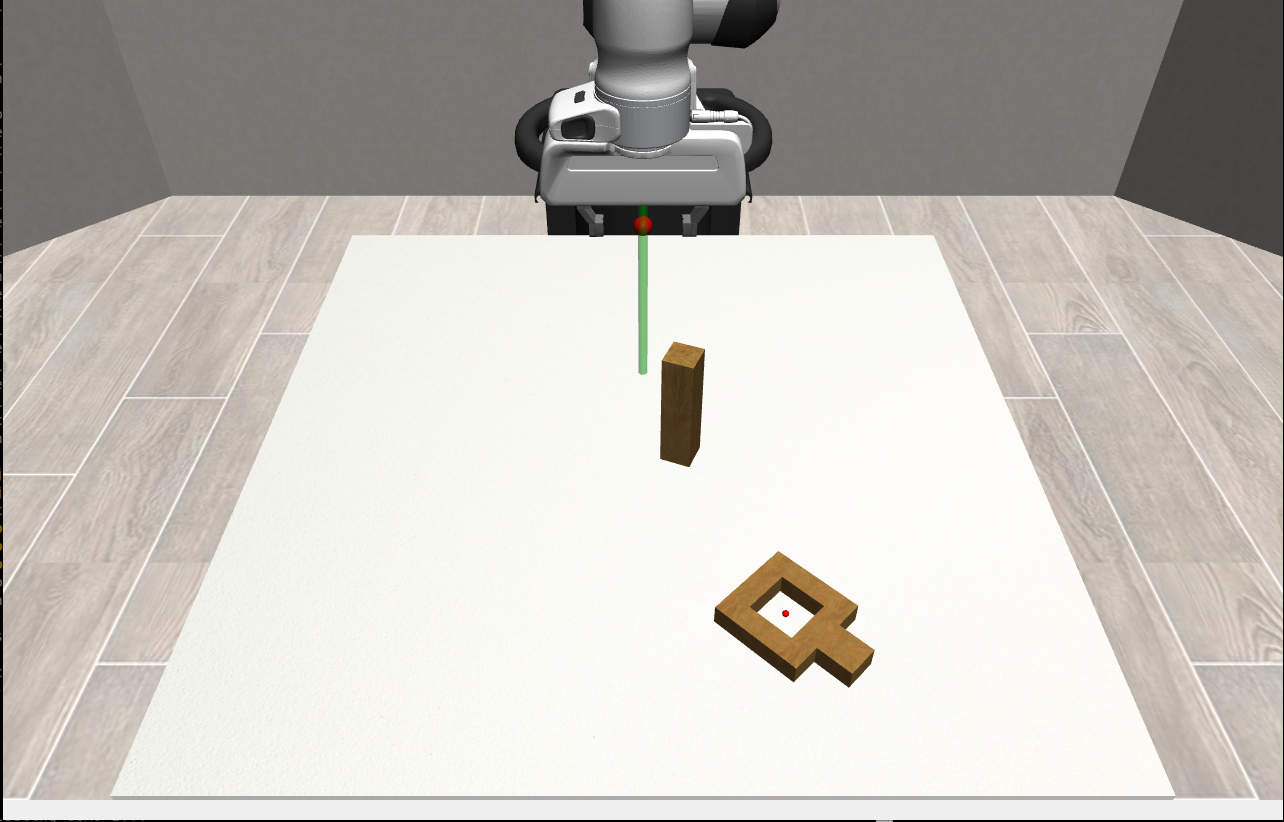}}
    \caption{Initial states sampled from the $D_0$, $D_1$, and $D_2$ distributions for the Square task. Each distribution introduces increasing variability in the positions and orientations of the nut and peg, illustrating how task difficulty and generalization demands grow across the three settings.}
    \label{fig:SquareExample}
\end{figure}

Each task includes a default reset distribution (D0), used for collecting all source demonstrations, along with broader distributions (D1 and D2) that introduce increasing complexity for data generation and policy learning.
For example, in the Square task, D0 randomizes only the nut (Figure \ref{fig:SquareExample}a), D1 adds variability to the peg’s position (Figure \ref{fig:SquareExample}b), and D2 randomizes both objects across new workspace regions (Figure \ref{fig:SquareExample}c).


\subsection{Dynamic Dataset Generation}
\label{subsec:dynamic_generation}

As discussed previously, D-MG is capable of handling dynamic changes in the environment during trajectory synthesis.
Accordingly, we evaluate the method’s ability to generate successful demonstrations under dynamic and perturbed conditions, which more closely resemble the challenges encountered in real-world robotic manipulation.

To evaluate D-MG's resilience to non-deterministic environments, we introduce dynamic disturbances during the trajectory generation process.
Specifically, we perturb the positions of target objects at runtime — during a segment of the Pick subtask — by applying randomized displacements to their spatial coordinates.

\begin{figure}[h]
    \centering
    \subfigure[\centering MugCleanup - Initial configuration]{\includegraphics[width=8.0cm]{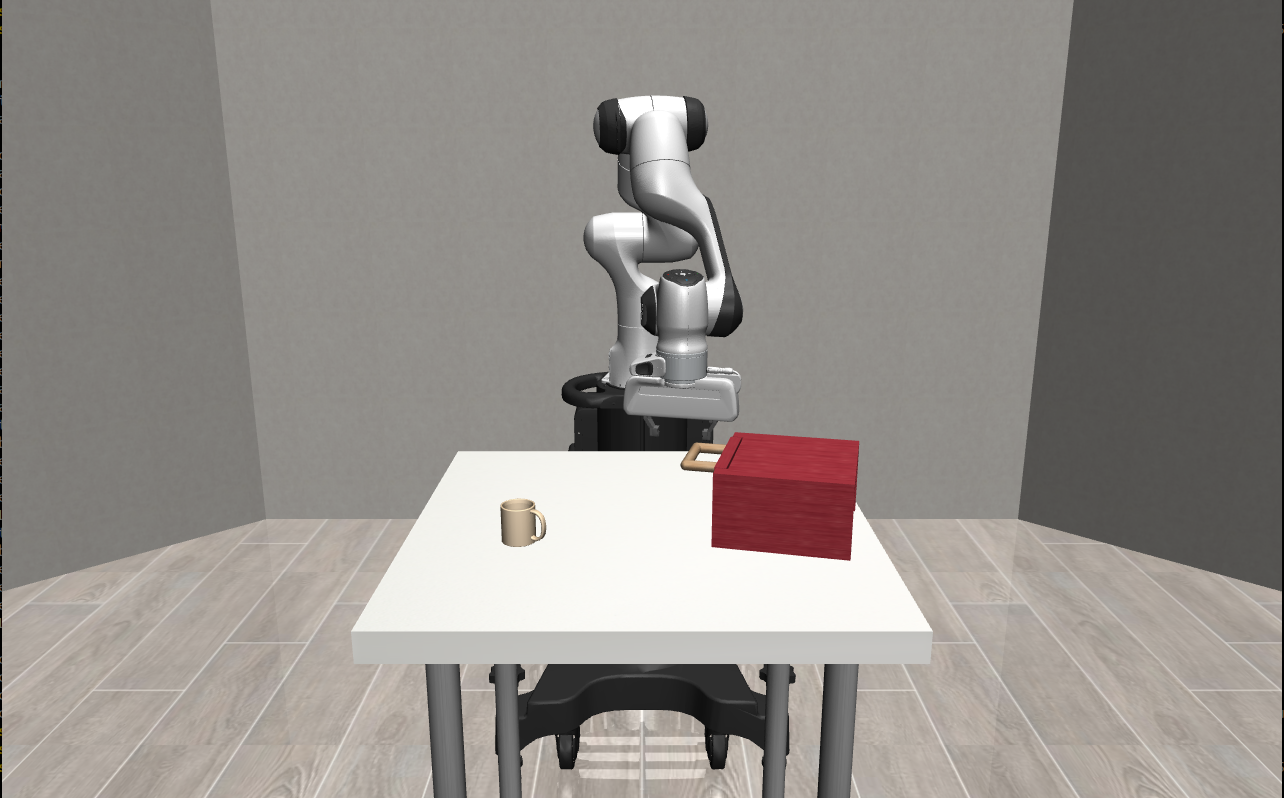}}
    \subfigure[\centering MugCleanup - Modified mug's position]{\includegraphics[width=8.0cm]{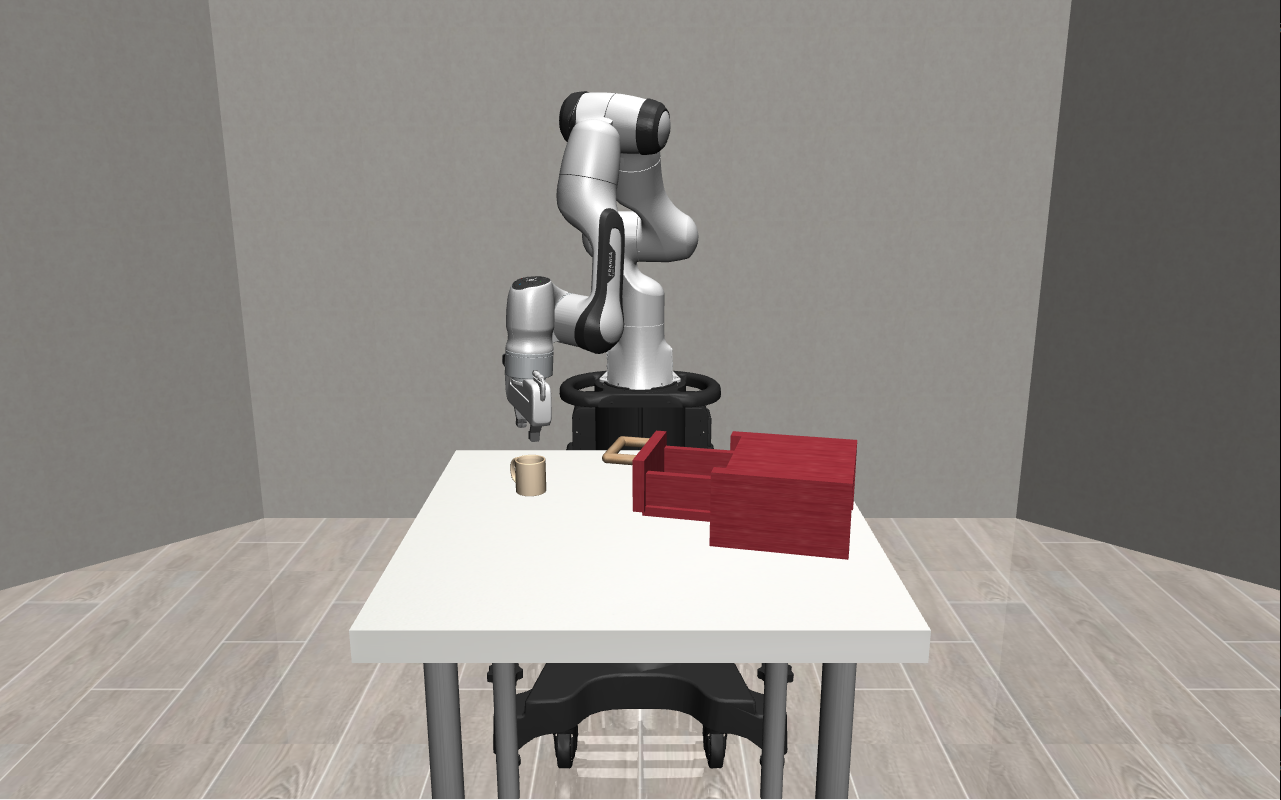}}\\
    \subfigure[\centering MugCleanup - Placing mug inside drawer]{\includegraphics[width=8.0cm]{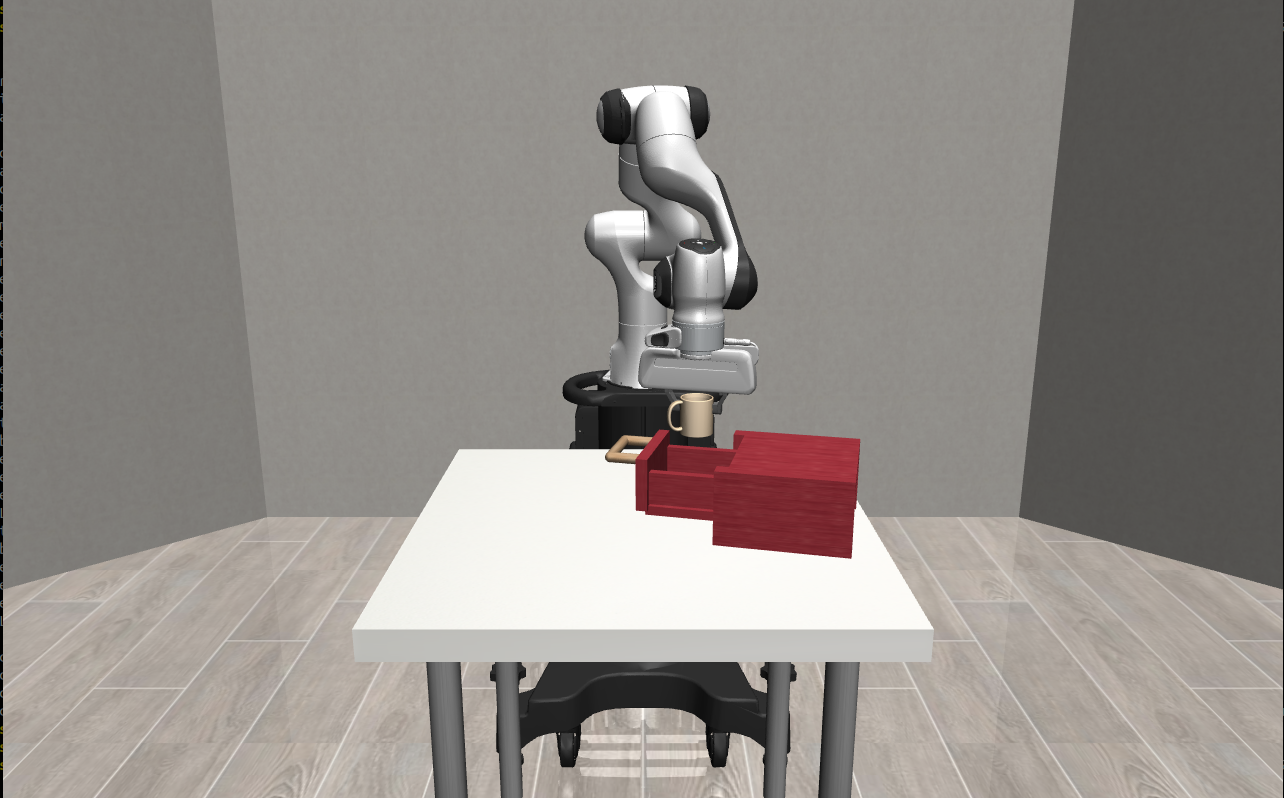}}
    \subfigure[\centering MugCleanup - Closing drawer]{\includegraphics[width=8.0cm]{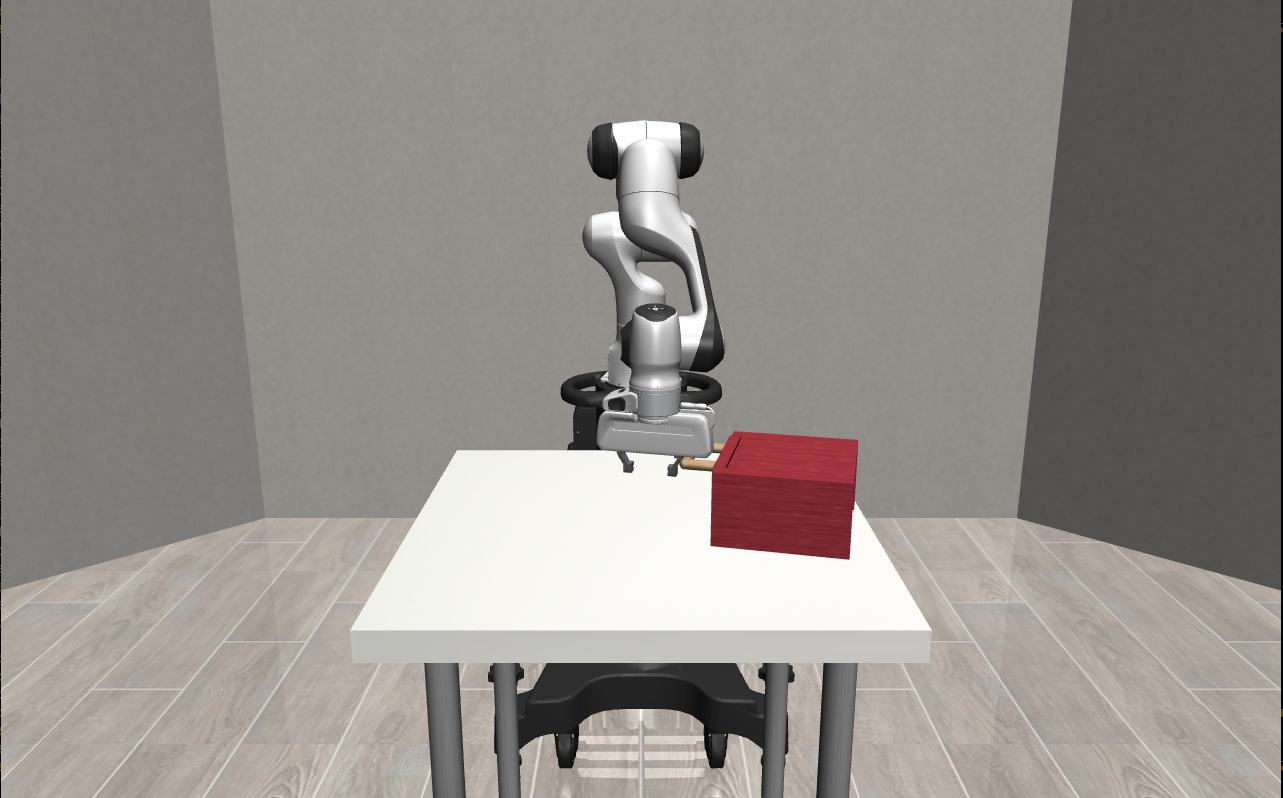}}
    \caption{\textbf{Dynamic Dataset Generation.} Example illustrating how DynaMimicGen (D-MG) introduces controlled perturbations to object positions during task execution.
    By continuously sensing the environment and adapting trajectories in response to these changes, D-MG generates diverse demonstrations that go beyond the original human demonstration, capturing a wide range of behaviors. This dynamic variation enhances the dataset’s richness, ultimately supporting the training of more robust and generalizable policies for robotic manipulation tasks.}
    \label{fig:dyn_env}
\end{figure}

As illustrated in Figure \ref{fig:dyn_env}, the \textit{MugCleanup} task demonstrates the robot’s ability to adapt to dynamic environmental changes.
Initially, the mug is placed in a specific location on the table Figure \ref{fig:dyn_env}a).
Subsequently, its position is altered (Figure~\ref{fig:dyn_env}b), requiring the robot to adjust its motion accordingly.
Leveraging the flexibility of Dynamic Movement Primitives (DMPs), the robot successfully adapts to the new object location, grasps the mug, places it inside the drawer (Figure \ref{fig:dyn_env}c), and completes the task by closing the drawer (Figure \ref{fig:dyn_env}d).
These perturbations force the data-generating policy to adjust its behavior mid-execution, simulating realistic execution-time uncertainties such as shifts in object location due to human interaction, sensor noise, or unmodeled physical forces.

This experimental setup has two goals: first, to evaluate whether D-MG can preserve demonstration quality when adapting online to deviations from the nominal trajectory; and second, to show that the resulting trajectories introduce enough variability to enable the training of robust and generalizable policies.

\subsection{Imitation learning algorithms}
\label{subsec:IL_algos}
For each task, a human operator collected a source dataset consisting of $N \in {1, 2}$ demonstrations on the default task variant via teleoperation \cite{mandlekar2018roboturk, mandlekar2019scaling}.
D-MG was then used to synthesize $1000$ successful demonstrations per task variant from these human demonstrations.
Because the generation procedure is not guaranteed to succeed on every attempt, unsuccessful rollouts (\textit{i.e.}, trajectories that failed to complete the task) were discarded, and data generation continued until $1000$ successful trajectories were obtained for each variant.
This process also enabled computation of the Data Generation Rate (DGR) for comparison of DynaMimicGen against MimicGen.

For each task, we train agents using Diffusion Policy~\cite{chi2024diffusionpolicy} and Behavior Cloning with an RNN-based policy~\cite{torabi2018behavioral}.
Behavior Cloning agents are trained on the full set of $1000$ generated trajectories, using the default hyperparameter configuration from robomimic~\cite{mandlekar2021matters}.
In contrast, Diffusion Policy agents are trained on a randomly sampled subset of $200$ trajectories, following the convention established in the original Diffusion Policy work~\cite{chi2024diffusionpolicy}, which demonstrates that the method achieves strong performance on robotic manipulation tasks even with datasets of this size.

Following the evaluation protocol in MimicGen \cite{mandlekar2023mimicgen}, we report policy performance as the maximum success rate over all evaluation checkpoints and three random seeds (see Appendix~\ref{app:add_exp_details} for full training details).
This allows for a fair and direct comparison of the quality of datasets produced by D-MG and MG in terms of their effectiveness for training imitation-learning policies.

\section{Experiment Results}
\label{sec:exp_results}
We conduct a comprehensive series of experiments designed to:
(1) demonstrate the versatility of DynaMimicGen (D-MG) in generating high-quality demonstration data across a diverse set of manipulation tasks;
(2) show that D-MG achieves competitive or superior data generation rates compared to existing methods, outperforming MimicGen in quasi-static scenarios (Section~\ref{subsubsec:static_gen}) while uniquely enabling data generation in dynamic environments (Section~\ref{subsubsec:dyn_gen});
(3) validate that agents trained with Behavior Cloning and Diffusion Policy on datasets generated by D-MG under dynamic scene variations consistently exhibit higher policy performance across all tasks compared to MG (Section~\ref{subsubsec:DP_performances});
(4) provide ablation studies highlighting the impact of dynamic scene perturbations in the D-MG generation process (Section~\ref{subsubsec:ablation_study}); and
(5) confirm the practical effectiveness of our approach through real-world experiments conducted on a physical robotic platform (Section~\ref{subsec:real_robot_exps}).

\subsection{Data Generation Performances}
In this section, we analyze the ability of DynaMimicGen to generate high-quality datasets for imitation learning in robotic manipulation.
Specifically, we compare D-MG and MimicGen (MG) from \cite{mandlekar2023mimicgen} in terms of Data Generation Rates.

\subsubsection{Data Generation Rates in Quasi-Static Task Settings}
\label{subsubsec:static_gen}
We first compare DynaMimicGen (D-MG) with MimicGen (MG) \cite{mandlekar2023mimicgen}.

\begin{table}[h]
    \centering
    \scriptsize
    \begin{minipage}{\textwidth}
        \centering
        \begin{tabular}{l|ccc|ccc|ccr}
        \multicolumn{1}{c|}{\textbf{Task}} & \multicolumn{3}{c|}{\textbf{D0}} & \multicolumn{3}{c|}{\textbf{D1}} & \multicolumn{3}{c}{\textbf{D2}}\\
        \textbf{} & \textbf{D-MG1} & \textbf{D-MG2} & \textbf{MG10} & \textbf{D-MG1} & \textbf{D-MG2} & \textbf{MG10} & \textbf{D-MG1} & \textbf{D-MG2} & \textbf{MG10}\\
        \hline
        Stack & \color{green}{95.0\%} & 94,8\% & 94.3\% & \color{green}{95.2\%} & 95,4\% & 90.0\% & & & \\
        Square & 53.9\% & \color{green}{81.4\%} & 73.7\% & 52.9\% & \color{green}{82.8\%} & 48.9\% & 37.6\% & \color{green}{63.2\%} & 31.8\% \\
        MugCleanup & 83.7\% & \color{green}{84.0\%} & 29.5\% & \color{green}{65.0\%} & 64.7\% & 17.0\% & & 
        \end{tabular}
        \caption{\textbf{Data Generation Rates}. For each task that we generated data for, we report the data generation rate (DGR) — which is the success rate of the data generation process - for our method (D-MG) and MimicGen (MG).}
        \label{tab:DMGvsMG}
    \end{minipage}
\end{table}

Table \ref{tab:DMGvsMG} reports the Data Generation Rates (DGR) achieved by DynaMimicGen (D-MG) and MimicGen (MG) across a diverse set of robot manipulation tasks, spanning basic stacking (\textit{Stack}), contact-rich assembly (\textit{Square}), and long-horizon multi-stage interaction (\textit{MugCleanup}).
While MG requires ten human demonstrations per dataset, D-MG attains comparable or superior data generation performance using only one (D-MG1) or two demonstrations (D-MG2), substantially reducing the amount of human supervision required.
This demonstrates the ability of D-MG to scale high-quality demonstration synthesis from extremely sparse expert demonstrations.

On the contact-rich \textit{Square} task, D-MG initially lagged behind MG when provided with only a single demonstration.
However, incorporating a second demonstration and applying an orientation-based selection strategy improved robustness in object-alignment subtasks, ultimately enabling D-MG to surpass MG’s performance.
These results highlight the method’s ability to benefit from marginal increases in data while still operating in a low-demonstration regime.

D-MG shows particularly strong performance on tasks involving dynamic object interactions (\textit{Stack}) and extended procedural structure (\textit{MugCleanup}), where trajectory adaptability and temporal coherence are critical.
Leveraging Dynamic Movement Primitives (DMPs), D-MG generates smooth, object-centric motion plans that remain consistent under environmental variability, enabling high DGRs while drastically minimizing human demonstration cost.

\subsubsection{Data Generation Rates in Dynamic Task Settings}
\label{subsubsec:dyn_gen}

\begin{table}[h]
    \centering
    \scriptsize
    \begin{minipage}{\textwidth}
        \centering
        \begin{tabular}{l|ccc|ccc|ccr}
        \multicolumn{1}{c|}{\textbf{Task}} & \multicolumn{3}{c|}{\textbf{D0}} & \multicolumn{3}{c|}{\textbf{D1}} & \multicolumn{3}{c}{\textbf{D2}}\\
        \textbf{} & \textbf{D-MG1} & \textbf{D-MG2} & \textbf{MG10} & \textbf{D-MG1} & \textbf{D-MG2} & \textbf{MG10} & \textbf{D-MG1} & \textbf{D-MG2} & \textbf{MG10}\\
        \hline
        Stack & \color{green}{68.7\%} & 67.2\% & \textit{N/A} & \color{green}{64.7\%} & 63.7\% & \textit{N/A} & & & \\
        Square & 42.7\% & \color{green}{67.6\%} & \textit{N/A} & 38.4 & \color{green}{64.3\%} & \textit{N/A} & 23.7 & \color{green}{52.9\%} & \textit{N/A} \\
        MugCleanup & 70.4\% & \color{green}{72.5\%} & \textit{N/A} & \color{green}{55.9\%} & 54.7\% & \textit{N/A} & & 
        \end{tabular}
        \caption{\textbf{Data Generation Rates.} For each task, we report the Data Generation Rate (DGR)—the success rate of the synthetic data generation process—under dynamic task settings.
        Because MG does not support dynamic feedback or disturbance recovery, it cannot be applied in these scenarios.
        As a result, no direct comparison with existing baselines is feasible, making this evaluation uniquely reflective of D-MG’s capabilities in dynamic environments.}
        \label{tab:dyn_gen}
    \end{minipage}
\end{table}

We assess DynaMimicGen's ability to generate successful demonstrations in dynamic environments by measuring the Data Generation Rates (DGR), across all the selected representative tasks, as reported in Table \ref{tab:dyn_gen}.

MimicGen (MG) is not designed to handle dynamic disturbances or time-varying perturbations in the environment.
Its data generation process assumes static conditions during execution, and it lacks mechanisms for online correction or adaptive motion refinement when unexpected interactions occur.
As a result, MG cannot be applied in scenarios where dynamic feedback and disturbance recovery are essential.
Consequently, no direct comparison with existing approaches is possible in this setting.
This makes our evaluation unique and highlights a capability that, to the best of our knowledge, is not addressed by current imitation-based data generation frameworks.

As expected, introducing dynamic perturbations leads to a reduction in success rates compared to the static environment setting (see Section \ref{subsubsec:static_gen}).
This performance drop is most pronounced when disturbances occur near task-critical events, particularly in tasks that require fine-grained control over object orientation, such as Square.
Such sensitivity is consistent with the inherent difficulty of precise contact-rich manipulation under non-stationary conditions.

Despite these challenges, DynaMimicGen exhibits strong resilience, consistently generating successful demonstrations across diverse manipulation tasks and perturbation regimes.
Importantly, success rates remain high across the $D0$, $D1$, and $D2$ distributions, highlighting the method’s ability to maintain reliable performance even as scene variability increases.

These results underscore the robustness of D-MG in settings where task dynamics deviate from the original demonstrations.
This capability is particularly crucial for real-world deployment, where environmental disturbances, object motion, and unexpected interactions are unavoidable.
By sustaining high data-generation quality under such variability, D-MG supports scalable and reliable imitation-learning pipelines, enabling policy training that better reflects the complexities of open-world robotic manipulation.

Overall, these findings demonstrate that D-MG is not only effective in scaling from minimal demonstrations but is also capable of sustaining high-quality data generation under dynamic and uncertain execution scenarios, making it a practical and versatile tool for real-world robotic learning applications.

\subsection{Imitation Learning Training Performances}
To rigorously evaluate the effectiveness of DynaMimicGen (D-MG) in enabling policy learning, we train agents using image-based Diffusion Policy (DP) and Behavior Cloning (BC) (see Appendix \ref{app:low_dim_results} for low-dim results) on the representative manipulation tasks — \textit{Stack}, \textit{Square}, and \textit{MugCleanup}.

\subsubsection{Agents Performance}
\label{subsubsec:DP_performances}

\begin{table}[h]
    \centering
    \scriptsize
    \begin{minipage}{1.0\textwidth}
        \centering
        \begin{tabular}{c|cc|cc}
            \toprule
            \textbf{Task} & \multicolumn{2}{c|}{\textbf{Diffusion Policy}} & \multicolumn{2}{c}{\textbf{Behavior Cloning}}\\
            \cmidrule(lr){2-3} \cmidrule(lr){4-5}
             & \textbf{D-MG} & \textbf{MG} & \textbf{D-MG} & \textbf{MG}\\
            \midrule
            \textbf{Stack} & & & & \\
            Source & $1.33 \pm 1.89$ & $1.33 \pm 1.89$ & $ 1.33 \pm 0.94$ & $0.67 \pm 0.94$ \\
            D0 & \textcolor{green}{\textbf{83.33 $\pm$ 3.40}} & $74.00 \pm 3.27$ & \textcolor{green}{\textbf{$76.00 \pm 3.27$}} & $72.67 \pm 1.89$ \\
            D1 & \textcolor{green}{\textbf{70.67 $\pm$ 3.40}} & $65.33 \pm 3.77$ & \textcolor{green}{\textbf{$71.33 \pm 0.94$}} & $58.00 \pm 1.63$ \\
            \hline
            \textbf{Square} & & & & \\
            Source & $3.33 \pm 2.49$ & $2.67 \pm 0.94$ & $2.00 \pm 0.00 $ & $2.00 \pm 1.63$ \\
            D0 & \textcolor{green}{\textbf{86.67 $\pm$ 0.90}} & $75.33 \pm 5.74$ & \textcolor{green}{\textbf{$97.33 \pm 0.94$}} & $88.67 \pm 0.94$ \\
            D1 & \textcolor{green}{\textbf{46.67 $\pm$ 4.71}} & $24.67 \pm 6.80$ & \textcolor{green}{\textbf{$50.67 \pm 0.94$}} & $44.00 \pm 2.83$ \\
            D2 & \textcolor{green}{\textbf{23.33 $\pm$ 2.50}} & $12.00 \pm 3.27$ & \textcolor{green}{\textbf{$37.33 \pm 2.49$}} & $34.00 \pm 4.32$ \\
            \hline
            \textbf{MugCleanup} & & & & \\
            Source & $2.00 \pm 1.63$ & $3.33 \pm 2.49$ & $0.67 \pm 0.94$ & $1.33 \pm 0.94$ \\
            D0 & \textcolor{green}{\textbf{90.00 $\pm$ 1.63}} & $61.33 \pm 9.27$ & \textcolor{green}{\textbf{$79.33 \pm 0.94$}} & $65.33 \pm 3.77$ \\
            D1 & \textcolor{green}{\textbf{60.67 $\pm$ 4.99}} & $38.00 \pm 2.83$ & \textcolor{green}{\textbf{$58.67 \pm 1.89$}} & $31.33 \pm 0.94$ \\
            \bottomrule
        \end{tabular}
        \caption{Performance comparison of image-based agents trained using Diffusion Policy and Behavior Cloning on datasets generated by DynaMimicGen (\textbf{D-MG}) versus those produced by the MimicGen (\textbf{MG}) baseline. 
        The table reports success rates across multiple tasks and data distributions, illustrating the extent to which D-MG’s dynamically generated demonstrations improve downstream policy learning.
        Overall, the results highlight the advantages of D-MG in producing higher-quality and more diverse training trajectories, leading to more robust and effective manipulation policies compared to MG.}
        \label{tab:results_image}
    \end{minipage}
\end{table}
We evaluate DynaMimicGen (D-MG) by training image-based agents using Diffusion Policy (DP) and Behavior Cloning (BC) algorithms on datasets synthesized from minimal human demonstrations.
This setting directly reflects the primary goal of D-MG: scaling one or two human demonstrations into large, diverse datasets suitable for imitation learning.
DP agents are trained for $2000$ epochs and BC agents for $600$ epochs, each across three random seeds, with the best-performing checkpoint from each run selected for evaluation (see Appendix~\ref{app:add_exp_details}).

For \textit{Stack} and \textit{MugCleanup}, only one demonstration is provided as input; for \textit{Square}, two demonstrations are used, as an additional example notably improves data generation robustness for this contact-rich task (Table \ref{tab:DMGvsMG}).
We then compare policies trained exclusively on source demonstrations to those trained on D-MG generated datasets.

Table~\ref{tab:results_image} demonstrates that datasets generated with DynaMimicGen (D-MG) lead to substantial improvements in downstream policy performance.
For example, in the \textit{Stack} task, the success rate increases dramatically from just $1.33\%$ when training on the original demonstration to $83.33\%$ and $70.67\%$ on the $D0$ and $D1$ data distributions, respectively, when using datasets generated with DynaMimicGen (D-MG).
This demonstrates D-MG’s ability to amplify limited supervision while preserving task structure. 
Similar improvements are observed in the other tasks, confirming the method’s capacity to capture multi-stage behavior.

Moreover, datasets generated with dynamic scene perturbations consistently produce the most effective policies across all tasks and data distributions.
Both Diffusion Policy and Behavior Cloning agents benefit from the richer variability and higher-quality demonstrations synthesized by D-MG, achieving higher and more stable success rates than with MimicGen (MG).
For instance, in the \textit{Square} task, DynaMimicGen (D-MG) achieves notable performance improvements over MG—raising success rates from $75.33\%$ to $86.67\%$ for Diffusion Policy agents and from $88.67\%$ to $97.33\%$ for Behavior Cloning agents in the $D0$ distribution.
This trend holds consistently across all evaluated tasks and data distributions, confirming that D-MG’s dynamic data generation strategy—which introduces controlled perturbations and adaptive trajectory refinements—produces more diverse and informative demonstrations.
As a result, policies trained on D-MG data generalize more effectively and perform more robustly, underscoring D-MG’s capability to generate high-quality, diverse, and policy-relevant training datasets for robotic manipulation.

While MimicGen requires ten human demonstrations to generate datasets with sufficient variability for training generalizable policies, D-MG achieves equal or superior performance using only one or two demonstrations. 
This improvement stems from D-MG's ability to introduce structured variability during trajectory generation by actively responding to dynamic perturbations in the environment, rather than relying solely on diversity present in the initial demonstrations.

Consequently, D-MG not only reduces the human data collection burden but also yields richer, more adaptable training datasets that better prepare learned policies to operate under real-world variability and disturbance conditions.

\subsubsection{Ablation Study}
\label{subsubsec:ablation_study}
This section investigates the role of dynamic scene variations in enhancing data quality during demonstration generation with DynaMimicGen (D-MG).
As discussed in previous sections, D-MG achieves significantly higher policy performance compared to MimicGen (MG) while requiring substantially fewer human demonstrations—only one for the \textit{Stack} and \textit{MugCleanup} tasks, and two for \textit{Square}.
In contrast, MG typically requires around ten demonstrations to produce datasets with sufficient variability to train generalizable policies.

The key advantage of D-MG lies in its ability to introduce structured variability during trajectory generation by actively responding to dynamic perturbations in the environment.
Instead of relying solely on the inherent diversity of the initial demonstrations, D-MG enriches the data through adaptive adjustments to motion trajectories—resulting in demonstrations that better capture the range of conditions encountered in real-world manipulation.
This dynamic data generation process ensures that the dataset contains diverse yet task-consistent trajectories, providing the variability required for robust policy learning.

In the absence of such dynamic perturbations, the generated trajectories tend to follow nearly identical motion patterns, differing only in their goal configurations.
This lack of variation limits the policy’s ability to generalize and adapt to unseen scenarios.
As shown in Table~\ref{tab:ablation_image}, datasets generated without dynamic scene variations yield markedly lower success rates across all evaluated tasks and data distributions.
In contrast, incorporating dynamic perturbations through D-MG leads to richer datasets and consistently higher policy performance, confirming the necessity of environmental variability during data synthesis.

\begin{table}[h]
    \centering
    \scriptsize
    \begin{minipage}{1.0\textwidth}
        \centering
        \begin{tabular}{c|cc|cc}
            \toprule
            \textbf{Task} & \multicolumn{2}{c|}{\textbf{Diffusion Policy}} & \multicolumn{2}{c}{\textbf{Behavior Cloning}}\\
            \cmidrule(lr){2-3} \cmidrule(lr){4-5}
             & \textbf{D-MG} & \textbf{MG} & \textbf{D-MG} & \textbf{MG}\\
            \midrule
            \textbf{Stack} & & & & \\
            Source & $0.67 \pm 0.94$ & $1.33 \pm 1.89$ & $ 0.00 \pm 0.00$ & $0.67 \pm 0.94$ \\
            D0 & $45.33 \pm 4.99$ & \textcolor{green}{\textbf{$74.00 \pm 3.27$}} & $32.00 \pm 4.32$ & \textcolor{green}{\textbf{$72.67 \pm 1.89$}} \\
            D1 & $28.00 \pm 4.32$ & \textcolor{green}{\textbf{$65.33 \pm 3.77$}} & $27.33 \pm 2.49$ & \textcolor{green}{\textbf{$58.00 \pm 1.63$}} \\
            \hline
            \textbf{Square} & & & & \\
            Source & $2.00 \pm 1.63$ & $2.67 \pm 0.94$ & $0.67 \pm 0.94 $ & $2.00 \pm 1.63$ \\
            D0 & \textcolor{green}{\textbf{$76.67 \pm 5.70$}} & $75.33 \pm 5.74$ & \textcolor{green}{\textbf{$94.00 \pm 1.63$}} & $88.67 \pm 0.94$ \\
            D1 & $11.33 \pm 3.40$ & \textcolor{green}{\textbf{$24.67 \pm 6.80$}} & $16.00 \pm 3.27$ & \textcolor{green}{\textbf{$44.00 \pm 2.83$}} \\
            D2 & $4.00 \pm 2.83$  & \textcolor{green}{\textbf{$12.00 \pm 3.27$}} & $20.67 \pm 6.35$ & \textcolor{green}{\textbf{$34.00 \pm 4.32$}}  \\
            \hline
            \textbf{MugCleanup} & & & & \\
            Source & $0.67 \pm 0.94$ & $3.33 \pm 2.49$ & $0.00 \pm 0.00$ & $1.33 \pm 0.94$ \\
            D0 & \textcolor{green}{\textbf{$81.33 \pm 7.36$}} & $61.33 \pm 9.27$ & \textcolor{green}{\textbf{$78.00 \pm 1.63$}} & $65.33 \pm 3.77$ \\
            D1 & \textcolor{green}{\textbf{$62.67 \pm 7.72$}} & $38.00 \pm 2.83$ & \textcolor{green}{\textbf{$37.33 \pm 2.49$}} & $31.33 \pm 0.94$ \\
            \bottomrule
        \end{tabular}
        \caption{Ablation study: Performance comparison of image-based agents trained using Diffusion Policy and Behavior Cloning on datasets generated with and without the dynamic perturbation mechanism of DynaMimicGen (\textbf{D-MG}), alongside the MimicGen (\textbf{MG}) baseline.
        The results show that disabling D-MG’s dynamic scene variations leads to significantly reduced policy performance, indicating that trajectories generated without these perturbations lack sufficient variability to support robust imitation learning.
        These results demonstrate the critical role of controlled dynamic variations in generating diverse, informative demonstrations and enabling stronger manipulation policies.}
        \label{tab:ablation_image}
    \end{minipage}
\end{table}

Interestingly, this trend is not always as pronounced when examining results obtained with low-dimensional state observations, as shown in Table~\ref{tab:ablation_low_dim}.
For Diffusion Policy agents, this behavior can be explained by the fact that, unlike image-based models that must learn compact and meaningful representations from raw high-dimensional visual inputs, low-dimensional agents operate directly on structured and semantically rich features such as object positions, orientations, and gripper states.
Consequently, their learning process is inherently less dependent on the variability introduced by D-MG’s dynamic data generation, since much of the environmental complexity is already encoded in the input state space.
While D-MG still contributes to performance improvements in several cases, the relative magnitude of these gains is smaller compared to the image-based setting, where dynamic perturbations significantly enhance visual generalization and robustness.

In contrast, when considering Behavior Cloning agents, the importance of dynamic perturbations becomes evident.
Agents trained on D-MG datasets that include dynamic variations consistently outperform those trained on static or less diverse data.
However, when such variability is not introduced, Behavior Cloning agents trained on MimicGen (MG) data can sometimes achieve higher success rates, indicating that D-MG’s advantages rely on the presence of dynamic perturbations to effectively diversify the training trajectories.
These observations underscore that while D-MG’s benefits are more evident in high-dimensional, visually complex domains, its dynamic data generation mechanism remains crucial even in structured low-dimensional settings to ensure sufficient variability for robust policy learning.

\begin{table}[h]
    \centering
    \scriptsize
    \begin{minipage}{1.0\textwidth}
        \centering
        \begin{tabular}{c|cc|cc}
            \toprule
            \textbf{Task} & \multicolumn{2}{c|}{\textbf{Diffusion Policy}} & \multicolumn{2}{c}{\textbf{Behavior Cloning}}\\
            \cmidrule(lr){2-3} \cmidrule(lr){4-5}
             & \textbf{D-MG} & \textbf{MG} & \textbf{D-MG} & \textbf{MG}\\
            \midrule
            \textbf{Stack} & & & & \\
            Source & $2.67 \pm 0.94$ & $4.67 \pm 2.49$ & $2.67 \pm 0.94$ & $3.33 \pm 0.94$ \\
            D0 & \textcolor{green}{\textbf{$96.00 \pm 1.63$}} & $96.00 \pm 1.63$ & $97.33 \pm 0.94$ & \textcolor{green}{\textbf{$98.67 \pm 0.94$}} \\
            D1 & $92.00 \pm 1.63$ & \textcolor{green}{\textbf{$92.67 \pm 1.89$}} & $75.33 \pm 3.40$ & \textcolor{green}{\textbf{$98.67 \pm 0.94$}} \\
            \hline
            \textbf{Square} & & & & \\
            Source & $3.33 \pm 0.94$ & $2.67 \pm 0.94$ & $3.33 \pm 0.94$ & $5.33 \pm 2.49$ \\
            D0 & $86.67 \pm 2.49$ & \textcolor{green}{\textbf{$91.33 \pm 0.94$}} & $94.67 \pm 0.94$ & \textcolor{green}{\textbf{$98.00 \pm 1.63$}} \\
            D1 & \textcolor{green}{\textbf{$89.33 \pm 2.49$}} & $86.67 \pm 1.89$ & $69.33 \pm 2.49$ & \textcolor{green}{\textbf{$88.67 \pm 3.40$}} \\
            D2 & $76.67 \pm 3.40$ & \textcolor{green}{\textbf{$86.67 \pm 0.94$}} & $68.00 \pm 4.32$ & \textcolor{green}{\textbf{$85.33 \pm 4.11$}} \\
            \hline
            \textbf{MugCleanup} & & & & \\
            Source & $0.00 \pm 0.00$ & $4.00 \pm 1.63$ & $0.00 \pm 0.00$ & $2.00 \pm 0.00$ \\
            D0 & \textcolor{green}{\textbf{$94.67 \pm 2.49$}} & $94.67 \pm 2.49$ & $97.33 \pm 0.94$ & \textcolor{green}{\textbf{$98.67 \pm 0.94$}} \\
            D1 & $74.67 \pm 2.49$ & \textcolor{green}{\textbf{$82.67 \pm 2.49$}} & \textcolor{green}{\textbf{$96.00 \pm 0.00$}} & $95.33 \pm 0.94$ \\
            \bottomrule
        \end{tabular}
        \caption{Ablation study: Performance comparison of low-dim agents trained using Diffusion Policy and Behavior Cloning on datasets generated with and without the dynamic perturbation mechanism of DynaMimicGen (\textbf{D-MG}), alongside the MimicGen (\textbf{MG}) baseline..}
        \label{tab:ablation_low_dim}
    \end{minipage}
\end{table}

These findings highlight that while D-MG’s core advantages are most evident in high-dimensional, visually grounded tasks, its underlying principles—structured data diversification and adaptive perturbation—still contribute positively across observation modalities.
In essence, the lower impact of D-MG in low-dimensional settings reflects not a limitation of the method, but the reduced need for variability when the state representation already encodes key task-relevant information explicitly.

\subsection{Experimental Validation}
\label{subsec:real_robot_exps}

\begin{figure}[h]
    \centering
    \subfigure[\centering Lift - initial configuration]{
        \includegraphics[width=0.3\textwidth]{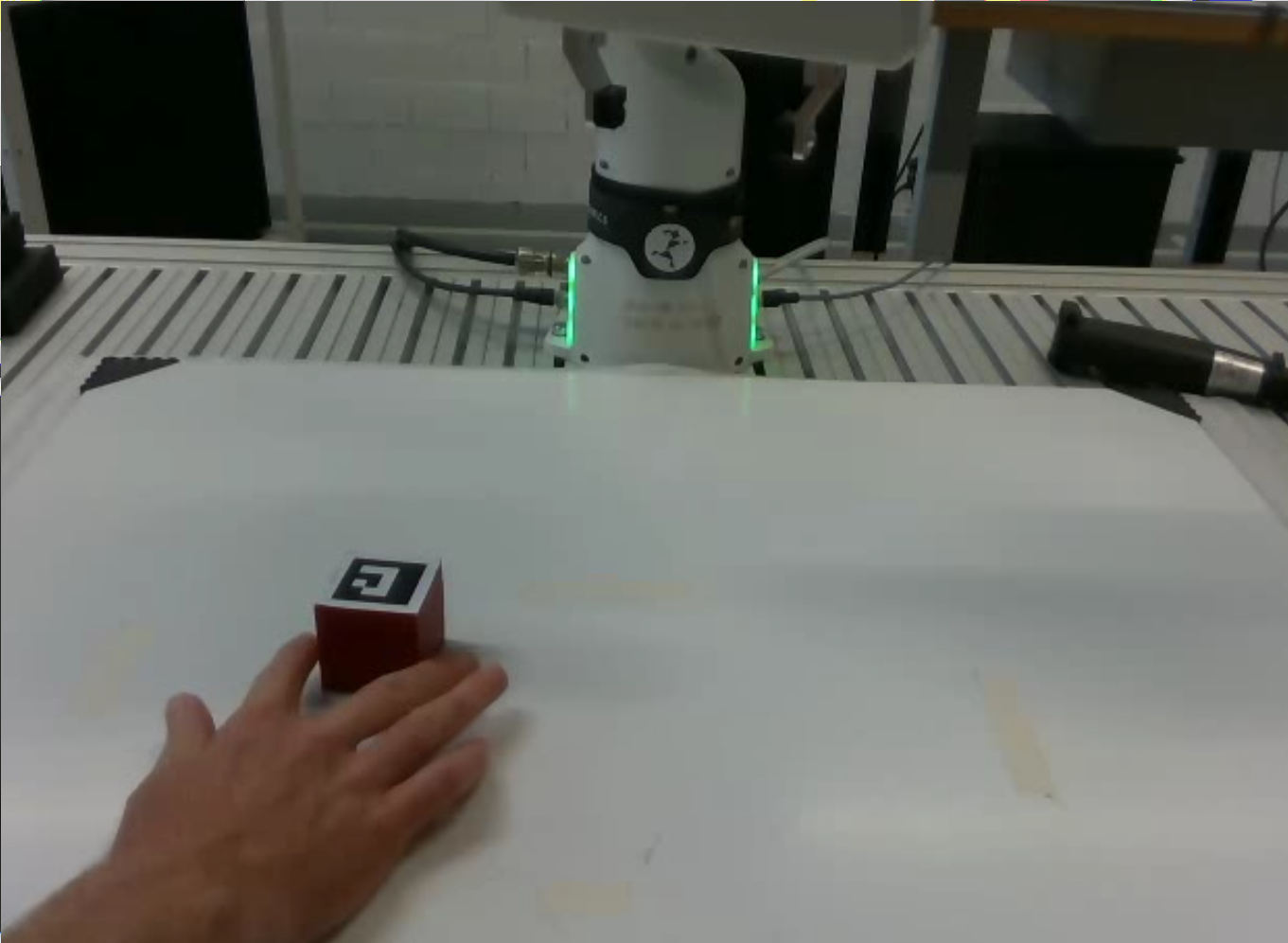}
    }
    \subfigure[\centering Lift - approach]{
        \includegraphics[width=0.3\textwidth]{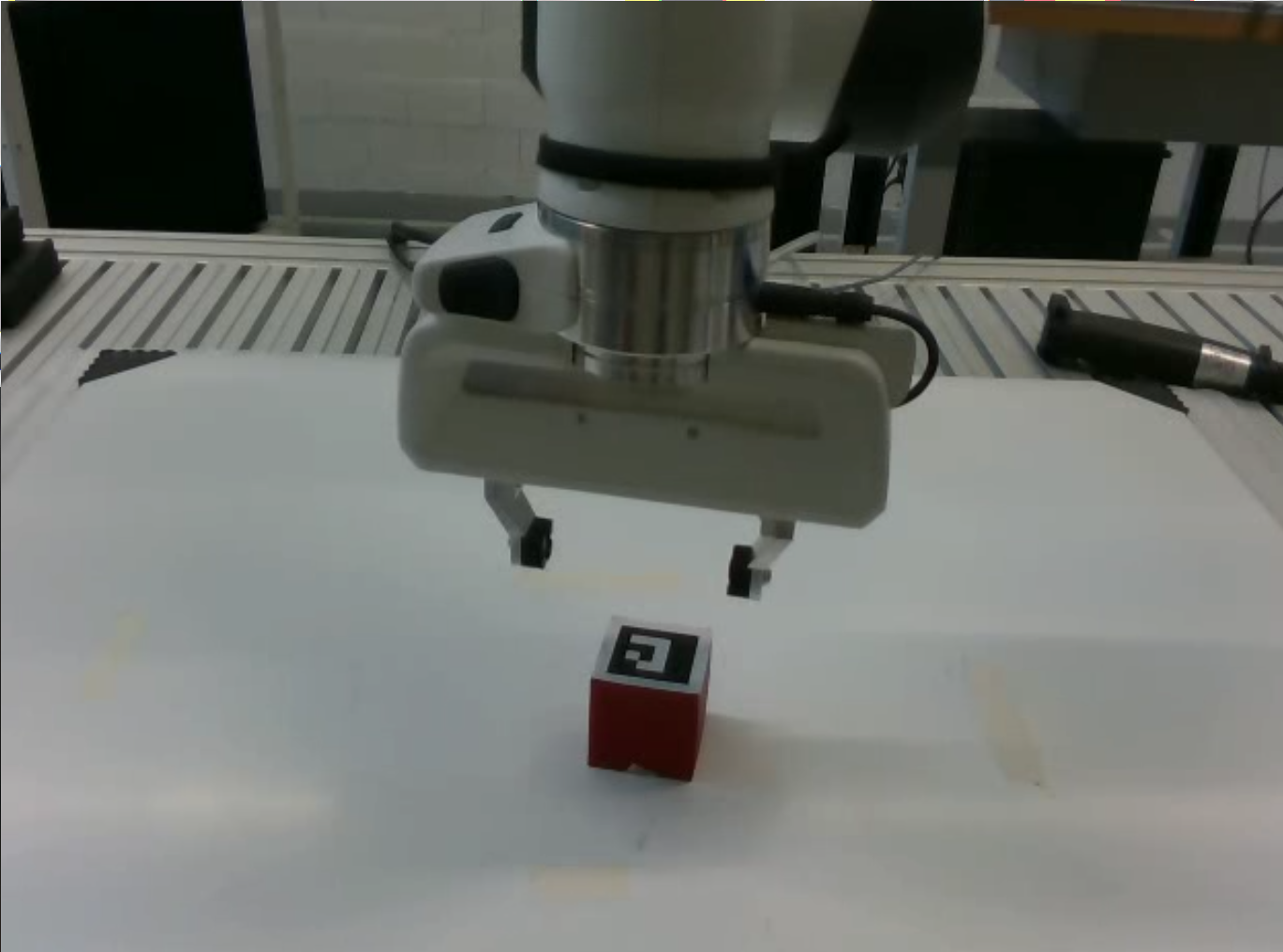}
    }
    \subfigure[\centering Lift - lifting]{
        \includegraphics[width=0.3\textwidth]{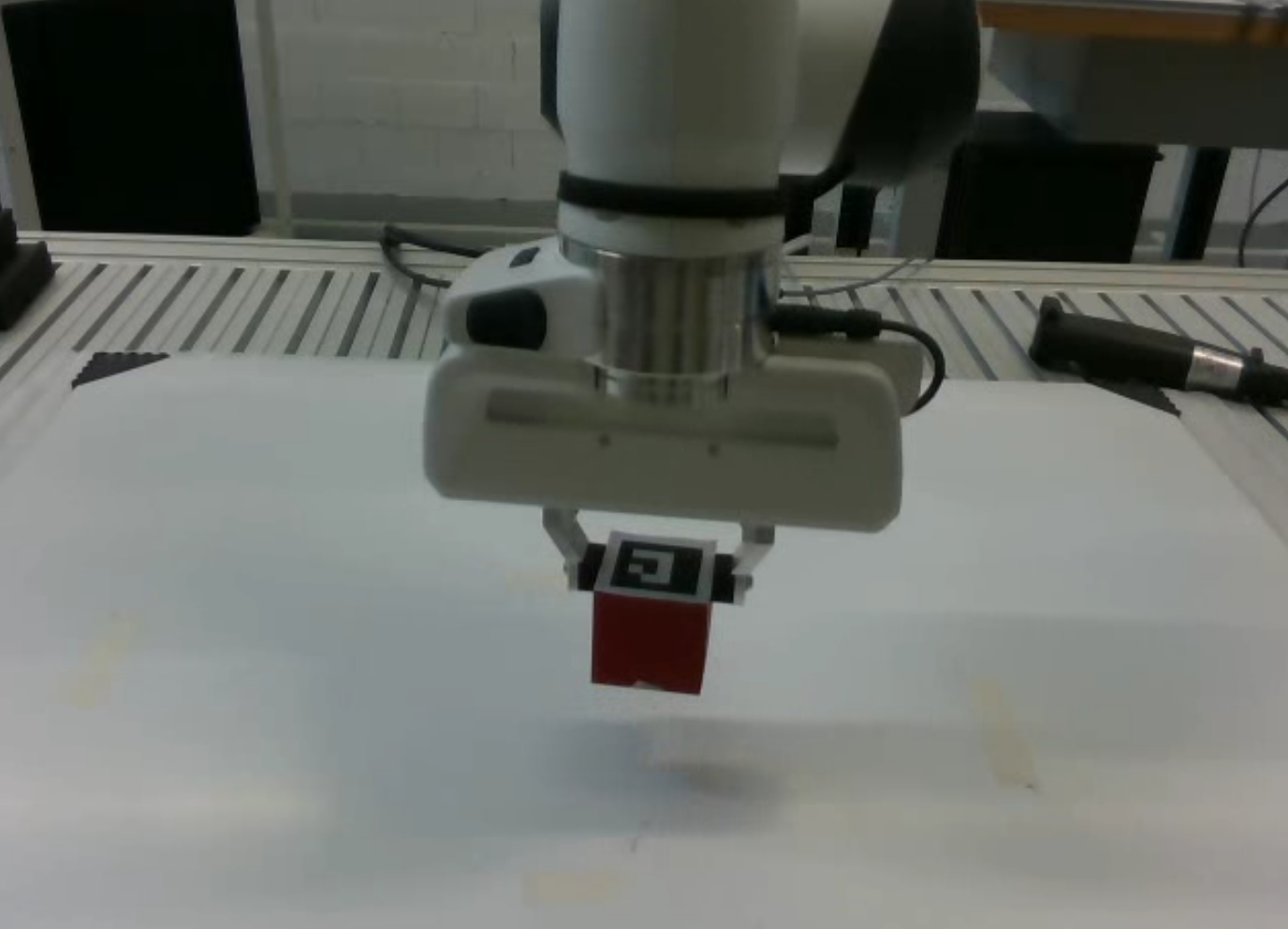}
    }\\
    \centering
    \subfigure[\centering Stack - initial configuration]{
        \includegraphics[width=0.3\textwidth]{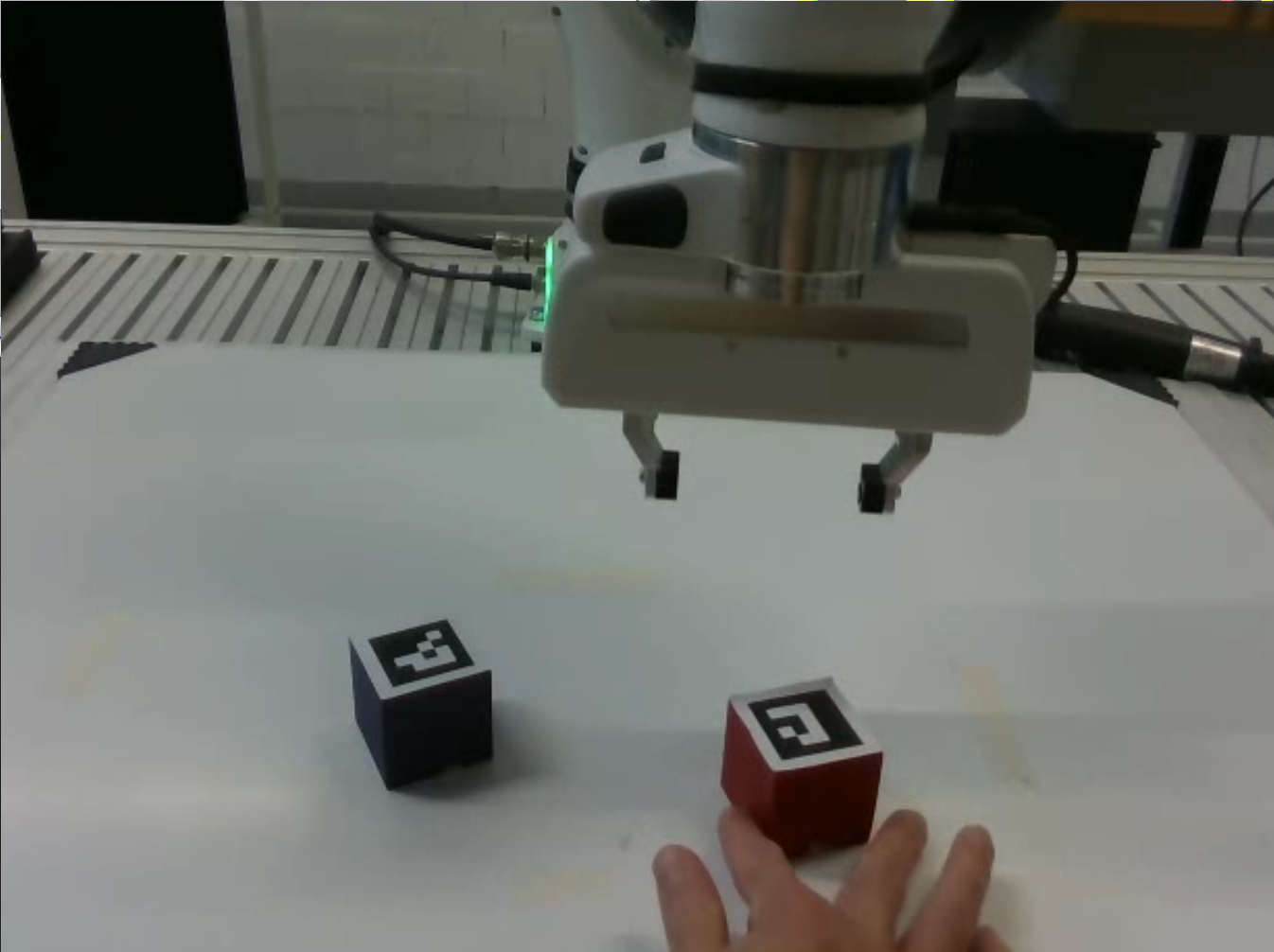}
    }
    \subfigure[\centering Stack - approach stacking]{
        \includegraphics[width=0.3\textwidth]{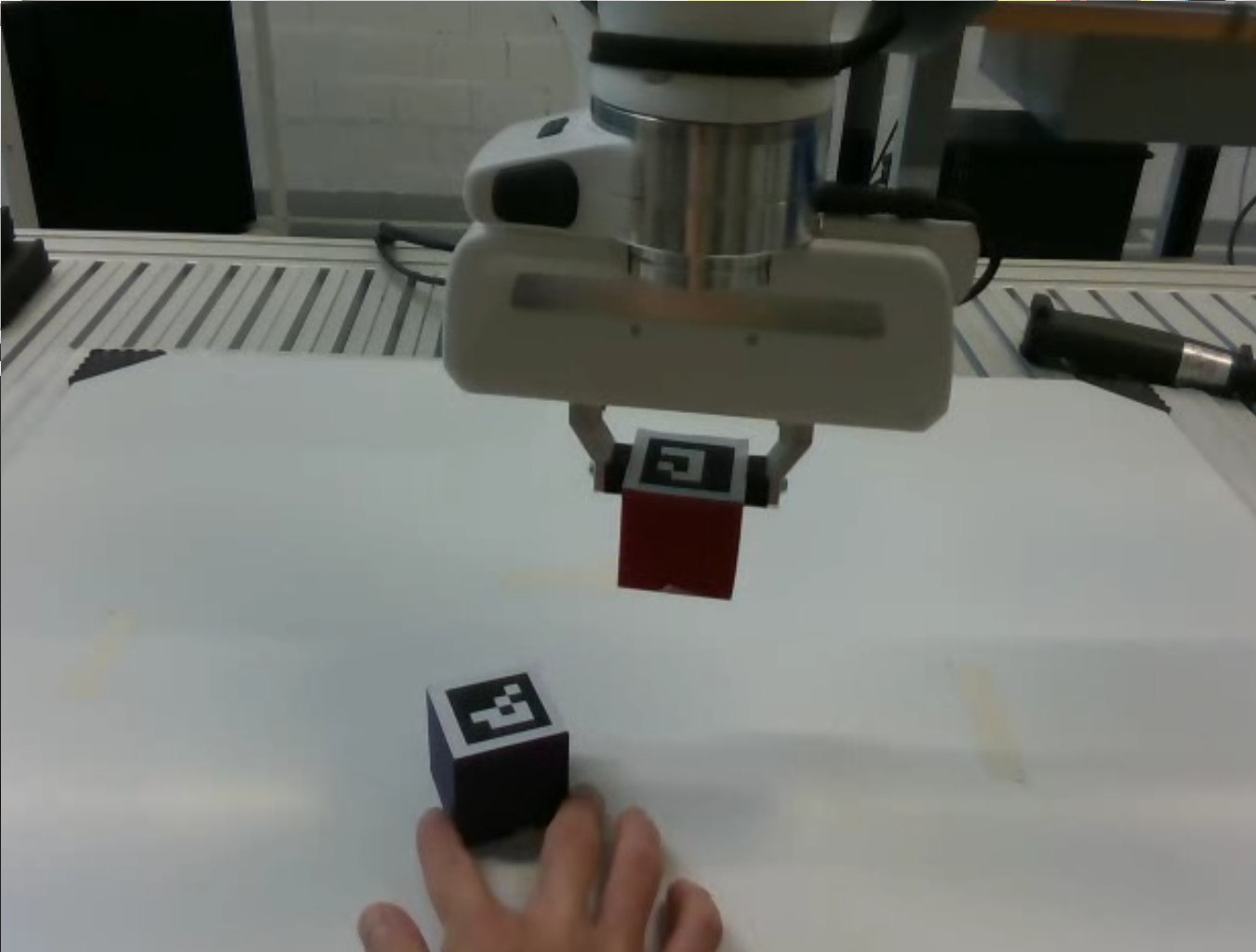}
    }
    \subfigure[\centering Stack - final configuration]{
        \includegraphics[width=0.3\textwidth]{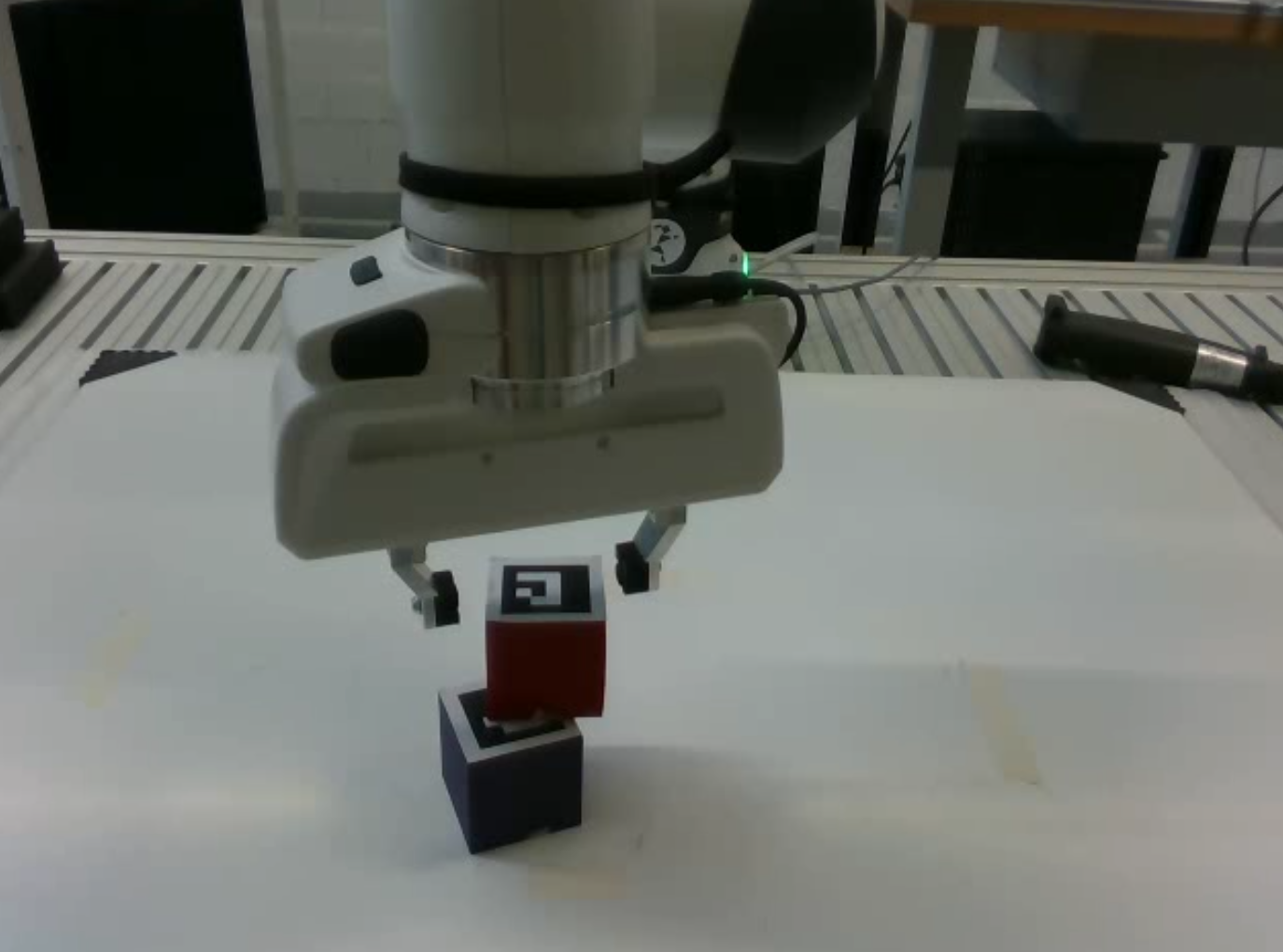}
    }\\
    \centering
    \subfigure[\centering Cleanup - Open box]{
        \includegraphics[width=0.3\textwidth]{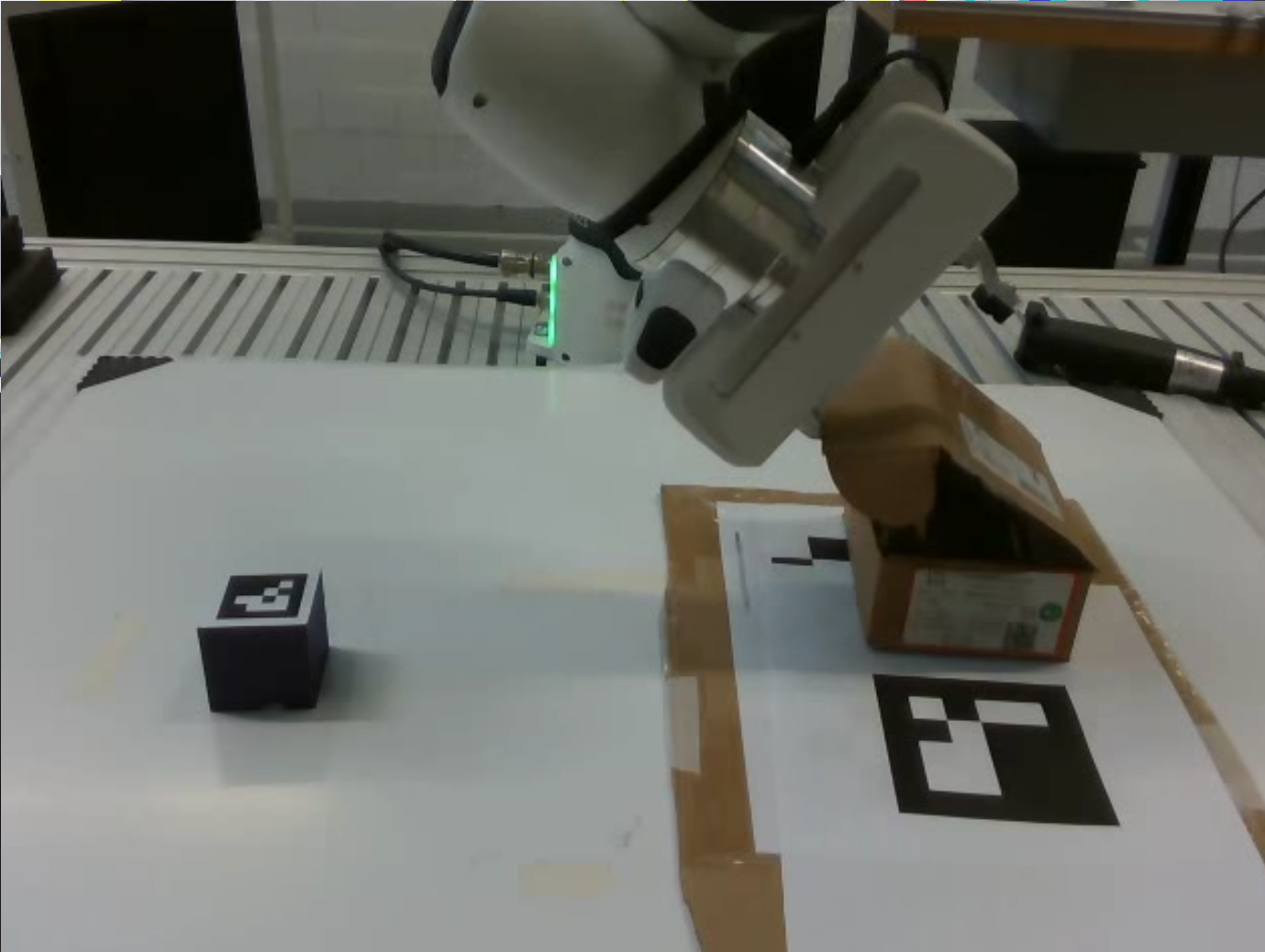}
    }
    \subfigure[\centering Cleanup - Object grasp]{
        \includegraphics[width=0.3\textwidth]{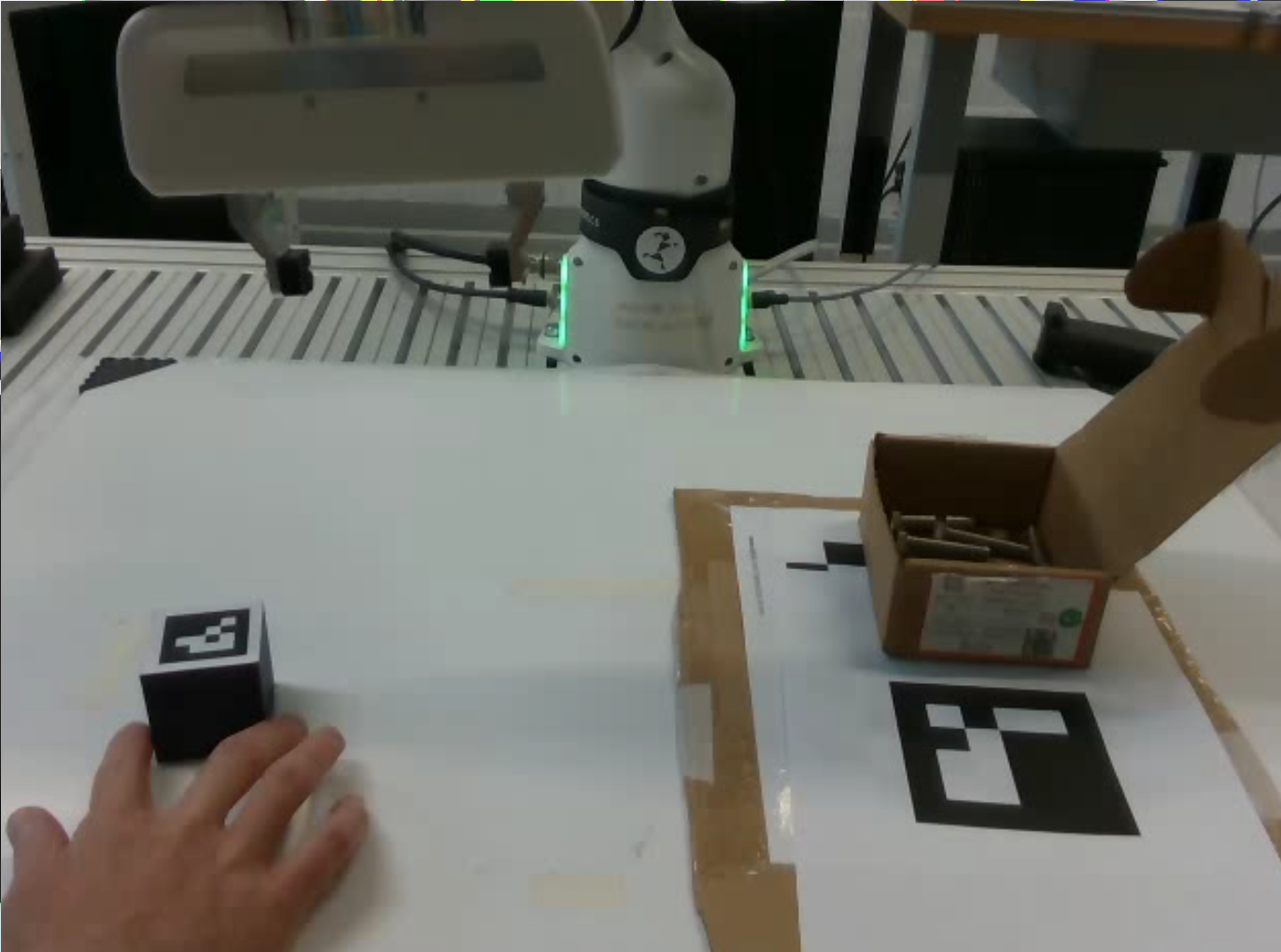}
    }
    \subfigure[\centering Cleanup - Object placed, closing box]{
        \includegraphics[width=0.3\textwidth]{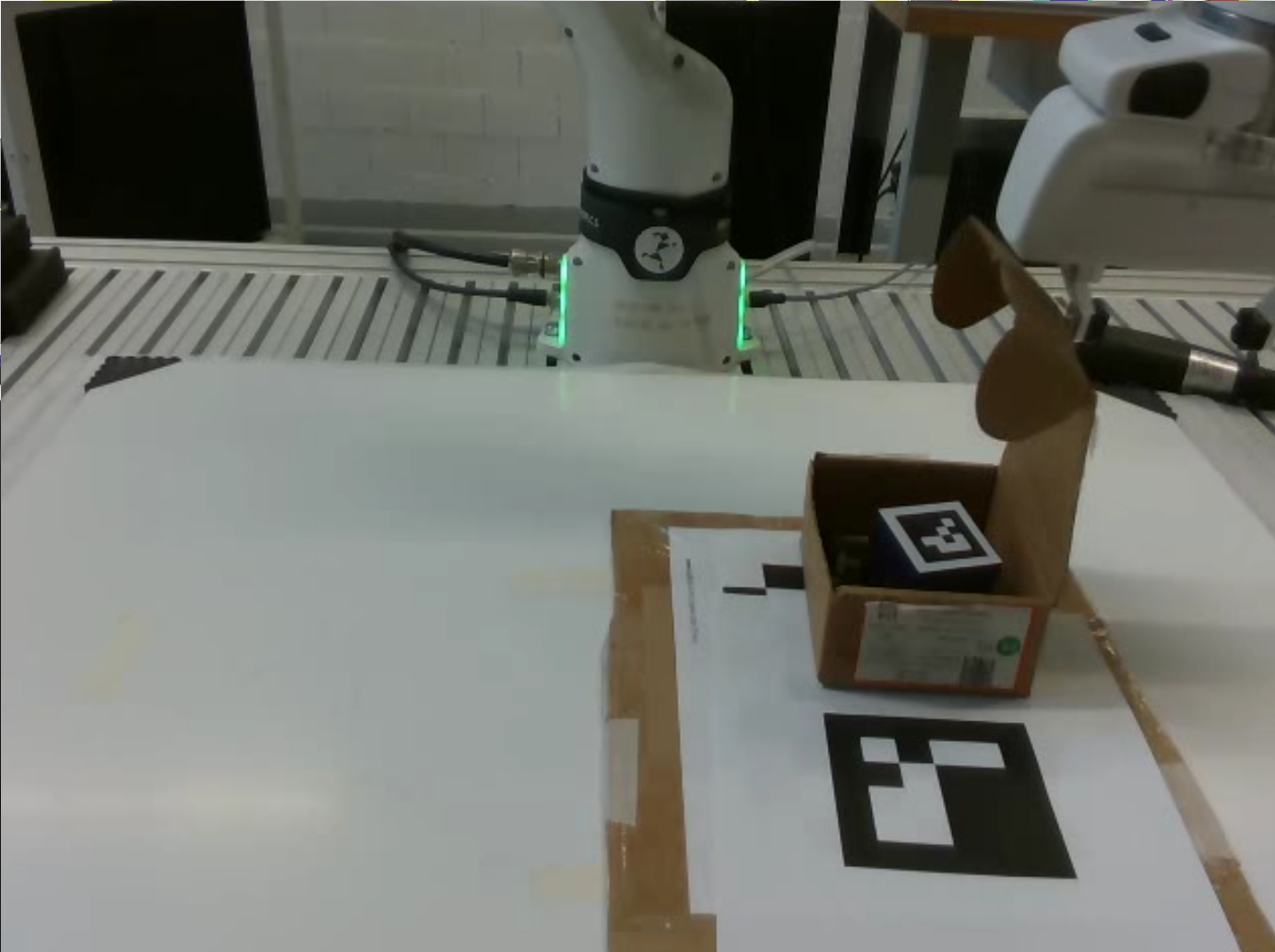}
    }
    \caption{\textbf{Real-World Subtasks.} Illustration of the three real-world manipulation tasks evaluated in our experiments: \textit{Lift}, \textit{Stack}, and \textit{Cleanup}. These tasks span different levels of complexity and interaction dynamics, providing a representative benchmark for assessing the robustness and generalization ability of policies trained using DynaMimicGen-generated datasets.}
    \label{fig:real_envs}
\end{figure}

We evaluated DynaMimicGen (D-MG) on a physical robotic platform — the Franka Emika Panda, a seven-degree-of-freedom manipulator — across three representative manipulation tasks: \textit{Lift} (Figure \ref{fig:real_envs} top-row), \textit{Stack} (Figure \ref{fig:real_envs} middle-row), and \textit{Cleanup} (Figure \ref{fig:real_envs}, bottom-row).
Additional implementation and setup details are provided in Appendix~\ref{app:real_robot_exp}.

The objective of these experiments is to validate D-MG’s practical effectiveness in real-world scenarios, specifically its ability to autonomously generate diverse, high-quality datasets that can be used to train deep learning–based control policies.
To this end, we employed Behavior Cloning (BC) as the test policy, chosen for its simplicity and direct interpretability in assessing demonstration quality.
By training BC agents solely on D-MG–generated trajectories and deploying them on the real robot, we verify that D-MG’s synthesized demonstrations transfer effectively to the physical domain, enabling policy learning without additional human input or manual retuning.
For each task, we measured both the trajectory generation success rate and the task execution success rate of the trained agents, as reported in Table~\ref{tab:real_exps}.

As in simulation, initial demonstrations were collected in static real-world scenes.
Subsequently, for each task, 50 new trajectories were generated while introducing dynamic scene variations, in which a human operator manually perturbed object positions during execution to simulate realistic disturbances.
In the \textit{Lift} and \textit{Stack} tasks, objects were displaced within a 25×50 cm area in front of the robot, whereas in the \textit{Cleanup} task, the square and box were restricted to smaller 25×20 cm and 10×5 cm regions, respectively.

\begin{table}[h]
    \centering
        \centering
        \begin{tabular}{lccc}
        \multicolumn{1}{c}{\textbf{Task}} & \textbf{Data Generation Success Rate} &  \textbf{BC Success Rate} \\
        \hline
        Lift & 94.0\%  & 86.0\%\\
        Stack & 88.0\% & 32.0\% \\
        Cleanup & 96.0\% & 26.0\% &\\
        \end{tabular}
        \caption{\textbf{Real-world evaluation.} Results obtained using the Franka Emika Panda robot arm on three dynamic manipulation tasks: Lift, Stack, and Cleanup. The table reports both the trajectory generation success rates using DynaMimicGen and the task success rates of policies trained with BC on the generated datasets.}
        \label{tab:real_exps}
\end{table}

We trained standard agents for 200 epochs and evaluated each over 50 rollouts.
As shown in Table~\ref{tab:real_exps}, D-MG achieved high trajectory generation success rates comparable to those observed in simulation, confirming the robustness of its generation process under physical dynamics.
Although policy success rates were relatively lower—primarily due to the limited dataset size—these results suggest that performance can be further improved by scaling the number of generated trajectories or employing more expressive policy learning architectures.

Overall, these findings confirm the feasibility and effectiveness of DynaMimicGen for semi-automatic dataset generation in dynamic, real-world conditions, demonstrating its potential as a scalable and hardware-agnostic tool for robotic learning from minimal human supervision.

\section{Conclusion}
\label{sec:conclusion}
\textbf{DynaMimicGen (D-MG)} introduces a scalable and adaptable framework for generating diverse, high-quality robot manipulation datasets from minimal human input.  
By integrating real-time trajectory adaptation mechanisms, D-MG enables robust and responsive data generation under dynamically changing environmental conditions.  
By continuously sensing the environment and adapting trajectories in response to these changes, D-MG generates diverse demonstrations that go beyond the original human demonstration, capturing a wide range of behaviors.
This dynamic variation enhances the dataset’s richness, ultimately supporting the training of more robust and generalizable policies for robotic manipulation tasks.  
This capability bridges the gap between static dataset generation pipelines and the demands of real-world robotic systems, where disturbances and environmental variability are inevitable.

Comprehensive experimental evaluations show that D-MG consistently achieves higher \textit{Data Generation Rates} (DGR) across a set of manipulation tasks and environment configurations, outperforming existing approaches such as \textit{MimicGen} not only in quasi-static scenes but also—crucially—in dynamic settings where real-time adaptation is required.  
A key factor behind these gains is D-MG’s ability to incorporate controlled dynamic perturbations during trajectory synthesis.

This dynamic variability directly translates into improved downstream policy performance: agents trained on D-MG datasets consistently achieve higher success rates than those trained on \textit{MimicGen}, exhibit stronger robustness to environmental perturbations, and generalize more effectively beyond the conditions seen during training.
These gains are not only evident in the main evaluation but are further reinforced by our ablation study, which shows that removing dynamic scene perturbations significantly degrades policy performance.
This demonstrates that the variability introduced by D-MG is a critical driver of the robustness and generalization capabilities of the learned policies.

Remarkably, D-MG attains these advantages using only one or two human demonstrations, whereas \textit{MimicGen} requires ten, underscoring D-MG’s ability to amplify minimal supervision into rich, diverse, and high-quality training data.

Moreover, real-world experiments on dynamically varying tasks—such as \textit{Lift}, \textit{Stack}, and \textit{Cleanup}—confirm the practical applicability of D-MG beyond simulation.  
These experiments validate its potential to accelerate data-driven robotic learning pipelines in physical environments, enabling effective imitation learning with minimal human involvement.\\
\textbf{Future Directions.}  
Future work will focus on extending D-MG to more complex and realistic dynamic scenarios, including obstacle-rich and multi-object environments where spatial reasoning and reactive adaptation are critical.  
Integrating explicit obstacle-awareness and physics-informed motion planning into the data generation process represents a promising direction for achieving safer and more generalizable dataset collection in unstructured real-world settings.  
Additionally, coupling D-MG with vision-language or foundation models could further enhance its ability to generalize across tasks, embodiments, and environments—paving the way toward scalable, autonomous data generation for next-generation robotic learning systems.\\
\textbf{Limitations.}  
Despite its versatility, D-MG is subject to several limitations that define its current scope and motivate future research.  

First, the framework assumes a known and fixed sequence of object-centric subtasks, where each subtask corresponds to a specific reference object.  
While this assumption simplifies task parsing and trajectory adaptation, it may constrain applicability to more flexible or unstructured task definitions in which subtask ordering is context-dependent.

Second, D-MG relies on accurate object pose estimates at every timestep during data generation.  
This requirement is essential for real-time adaptation but introduces a dependency on reliable perception systems—posing challenges in settings with occlusions, sensor noise, or complex object geometries.

Furthermore, D-MG currently supports only one reference object per subtask, limiting its capacity to model tasks involving coordinated motion relative to multiple objects (e.g., insertions or placements in cluttered environments).  
Extending the framework to multi-object dependencies remains an open research challenge.

The system is also currently restricted to single-arm manipulation, which simplifies coordination but prevents application to bimanual or dual-arm scenarios—common in more advanced manipulation settings.

Finally, although D-MG demonstrates strong performance in simulation, its deployment on physical platforms has been limited to a few representative tasks (\textit{Lift}, \textit{Stack}, and \textit{Cleanup}).  
Broader experimental validation across different robots and real-world conditions will be essential to fully assess its robustness and practicality.  
Addressing these limitations constitutes an important avenue for future work.  
Additional discussion and implementation details are provided in Appendix~\ref{app:limitations}.



\vspace{6pt}

\newpage

\appendix
\section[\appendixname~\thesection]{Overview}
\label{app:overview}
The Appendix contains the following content:
\begin{itemize}
    \item \textbf{Author Contributions} (Appendix \ref{app:authors_contrib}): list of each author’s contributions to the paper.
    \item \textbf{FAQ} (Appendix \ref{app:faq}): answers to some common questions.
    \item \textbf{Additional Experiment Details} (Appendix \ref{app:add_exp_details}): additional details on agent training procedures.
    \item \textbf{Low-Dim Results} (Appendix \ref{app:low_dim_results}): results of agents trained using low-dim state observations.
    \item \textbf{Real Robot Experimental Details}: (Appendix \ref{app:real_robot_exp}): additional details and discussion on the real robot experiments.
    \item \textbf{Limitations and Failure Modes}: (Appendix \ref{app:limitations}): description of limitations and failure modes of DynaMimicGen.
\end{itemize}

\newpage

\section[\appendixname~\thesection]{Author Contributions}
\label{app:authors_contrib}
\textbf{Vincenzo Pomponi} led the overall project, implemented the DynaMimicGen code, ran the experiments in the paper, and wrote the paper.

\textbf{Paolo Franceschi} developed the DynaMimicGen code used in real-world experiments, provided experiment ideas, and supervised the writing of the paper.

\textbf{Stefano Baraldo} provided project guidance, contributed experimental ideas, and supervised the writing of the paper.

\textbf{Oliver Avram} advised on the project.

\textbf{Loris Roveda} advised on the project, contributed experimental ideas, and supervised the writing of the paper.

\textbf{Luca Maria Gambardella} advised on the project.

\textbf{Anna Valente} advised on the project.

\newpage

\section[\appendixname~\thesection]{FAQ}
\label{app:faq}

\textbf{Why train policies with DynaMimicGen-generated datasets instead of directly deploying DynaMimicGen for task execution?}\\
While DynaMimicGen excels at generating successful demonstrations by adapting a single human demonstration to dynamic variations in the environment, it operates by leveraging one demonstration at a time.  
As a result, it lacks the capacity to merge information across multiple demonstrations, such as combining different strategies or variations in object shapes and interaction behaviors to achieve a more robust execution.  
In contrast, training a policy on a dataset generated with DynaMimicGen — particularly when using multiple human demonstrations as input — enables the learning system to generalize across diverse trajectories, effectively capturing richer task-relevant variations.  
This leads to policies that are more adaptable and resilient to environmental changes, object diversity, and execution noise, capabilities that go beyond what DynaMimicGen alone can achieve during direct trajectory execution.

\textbf{Are Data Generation Success Rates Correlated with Trained Agent Performance?}\\
While it may be intuitive to assume a correlation between data generation success rates and the performance of trained agents, our findings indicate that this relationship does not necessarily hold.
In several cases, datasets with relatively low generation success rates—thus requiring more time to collect 1000 successful trajectories—still yielded high agent performance after Diffusion Policy (DP) training.
Notably, in the \textit{MugCleanup} task under the $D_0$ distribution, DynaMimicGen achieved a data generation success rate of 55.9\%, while agents trained on the resulting datasets attained success rates of 90.00\% with image-based observations and 96.67\% with low-dimensional inputs in dynamic task settings.
This highlights both the efficiency of D-MG’s data generation process and the high utility of the synthesized trajectories for downstream policy learning across different observation modalities.
Moreover, this emphasizes the value of using replay-based mechanisms for data collection rather than deploying data generation policies directly, as done in prior works~\cite{wen2022you, di2022learning}.
The apparent disconnect stems from the fact that agent training via DP is based on a fixed number of successful demonstrations (200), irrespective of how many attempts were required to obtain them.

Consequently, even when DynaMimicGen encounters difficulty in consistently generating successful trajectories—particularly under challenging task conditions—the resulting curated datasets remain effective for training.
This highlights the robustness of DynaMimicGen: despite imperfect generation success rates in complex environments, once a sufficient number of successful demonstrations are obtained, DP can still produce policies with strong generalization and high success at deployment.

\textbf{Why comparing D-MG and MG using absolute end-effector poses?}\\
D-MG represents trajectories using Dynamic Movement Primitives (DMPs), which require the full sequence of absolute end-effector poses for training and online generalization to new object configurations. Consequently, our evaluation necessitates a controller capable of executing absolute Cartesian goals.

In contrast, MimicGen (MG) \cite{mandlekar2023mimicgen} reports results using delta end-effector actions.
These delta commands cannot be reliably converted into absolute poses because the PD controller used in RoboSuite, does not guarantee perfect convergence to each intermediate waypoint.
As a result, the executed end-effector poses deviate from the commanded deltas, making it impossible to accurately reconstruct the true absolute trajectory from delta-based control signals.
However, delta trajectories can be converted to absolute form by replaying them in simulation and recording the resulting end-effector states.

For a fair comparison, we therefore re-executed MG trajectories to obtain their corresponding absolute pose sequences, and re-trained and evaluated Diffusion Policy on these reconstructed datasets.
This procedure ensures consistency with D-MG’s absolute-trajectory formulation and enables a fair and meaningful comparison with MG.

\textbf{Why does MG exhibit different policy performance when trained on absolute end-effector poses compared to delta-based trajectories?}\\
When analyzing the results, we observed notable discrepancies between the performance metrics reported in the original MimicGen paper~\cite{mandlekar2023mimicgen} and those presented in Tables \ref{tab:results_image}, \ref{tab:results_low_dim} for Behavior Cloning (BC) agents trained using delta and absolute end-effector trajectories, respectively.
To investigate this, we conducted controlled experiments where two BC agents were trained with identical hyperparameters — one on datasets containing delta (relative) end-effector motions, and the other on datasets converted to absolute (global) end-effector poses derived from the same demonstrations.

Our findings indicate that the difference in policy performance arises primarily from the way the model learns to associate changes in the robot’s end-effector pose with state transitions in the environment.
In particular, when trained on image-based observations, policies benefit from the delta representation, as it provides a more direct and local correspondence between action outputs (i.e., motion deltas) and the visual changes observed in consecutive frames.
This relative encoding simplifies the learning problem by reducing the dependency on global spatial references, which can be difficult to infer from images alone.

Conversely, when trained on low-dimensional state representations that include precise object poses, the advantage of the delta formulation diminishes.
In this case, absolute pose information can be effectively leveraged by the policy to learn consistent mappings between global robot positions and object configurations.
Hence, the observed discrepancies between delta- and absolute-based policies reflect a fundamental difference in how spatial information is represented and exploited across observation modalities.

\newpage

\section[\appendixname~\thesection]{Additional Experiment Details}
\label{app:add_exp_details}
We describe details of how policies were trained via imitation learning.
Several design choices are the same as the robomimic study \cite{mandlekar2021matters}.

\textbf{Observation Spaces.}
Following the protocol established in RoboMimic \cite{mandlekar2021matters}, we train policies using two types of observation spaces: \textit{low-dim} and \textit{image}.
Both modalities include the robot end-effector pose and gripper finger positions.
The \textit{low-dim} setting additionally incorporates ground-truth object poses, while the \textit{image} setting provides visual inputs from two camera perspectives: a front-view camera and a wrist-mounted camera.
Since our method relies on absolute end-effector poses (as discussed in Assumption 1, Section~\ref{sec:problem}), and to ensure compatibility with the output range expected by Behavior Cloning and Diffusion Policy architectures, we normalize end-effector poses to the range $[-1, 1]$ during training.
These values are subsequently denormalized during evaluation.
All image-based experiments utilize RGB inputs at a resolution of $84 \times 84$ pixels.

\textbf{Training Hyperparameters.}
We employ BC-RNN using the default hyperparameter configuration from robomimic \cite{mandlekar2021matters} with the default hyperparameters reported in their study,with the exception of an increased learning rate (1e-3 instead of 1e-4) for policies trained on low-dim observations, as we found it to speed up policy convergence on large datasets.\\
For Diffusion Policy \cite{chi2024diffusionpolicy}, we use the following training settings: an initial learning rate of ($1e^{-4}$) decaying to $0.0$ as the number of epochs increases, $2000$ training epochs for both image-based and low-dimensional agents, $500$ warm-up steps, and a batch size of $16$ for image-based agents and $100$ for low-dimensional agents.

\textbf{Policy Evaluation.} As in \cite{mandlekar2021matters}, on simulation tasks, we evaluate policies using 50 rollouts per agent checkpoint during training, and report the maximum success rate achieved by each agent across 3 seeds.
On the real world tasks, due to the time-consuming nature of policy evaluation, we take the last policy checkpoint produced during training, and evaluate it over 50 episodes.

\textbf{Hardware.} Each data generation run and training run used a machine with an NVIDIA GeForce RTX 4090 GPU, 24 GB of memory, and 1 TB of disk space.
In certain cases, we batched multiple data generation runs and training runs on the same machine (usually 2 runs).
Real robot experiments were carried out on a machine running Ubuntu 22.04, with PREEMPT\_RT kernel, with 20 CPUs, 32GB of memory, and 1 TB of storage, and no dedicated GPU. The robotic platform is a Franka Research 3, controlled via the ROS2 interface.

\newpage

\section[\appendixname~\thesection]{Low-Dim Results}
\label{app:low_dim_results}
In the main text, our analysis primarily focused on policies trained using image-based observation spaces, as they represent a more realistic and challenging perception setting for visuomotor learning.
In this section, we extend the evaluation to agents trained with Diffusion Policy on low-dimensional (low-dim) state representations, which directly encode the robot and object poses rather than relying on raw pixel inputs.
This additional evaluation allows us to assess whether the benefits of DynaMimicGen (D-MG) hold across different observation modalities and to confirm that the generated datasets remain effective beyond visual-based policy learning.

\begin{table}[h]
    \centering
    \scriptsize
    \begin{minipage}{1.0\textwidth}
        \centering
        \begin{tabular}{c|cc|cc}
            \toprule
            \textbf{Task} & \multicolumn{2}{c|}{\textbf{Diffusion Policy}} & \multicolumn{2}{c}{\textbf{Behavior Cloning}}\\
            \cmidrule(lr){2-3} \cmidrule(lr){4-5}
             & \textbf{D-MG} & \textbf{MG} & \textbf{D-MG} & \textbf{MG}\\
            \midrule
            \textbf{Stack} & & & & \\
            Source & $2.00 \pm 0.00$ & $4.67 \pm 2.49$ & $4.00 \pm 2.83$ & $3.33 \pm 0.94$ \\
            D0 & \textcolor{green}{\textbf{$96.67 \pm 2.49$}} & $96.00 \pm 1.63$ & $98.00 \pm 0.00$ & \textcolor{green}{\textbf{$98.67 \pm 0.94$}} \\
            D1 & \textcolor{green}{\textbf{$92.67 \pm 1.89$}} & $92.67 \pm 1.89$ & \textcolor{green}{\textbf{$99.33 \pm 0.94$}} & $98.67 \pm 0.94$ \\
            \hline
            \textbf{Square} & & & & \\
            Source & $5.33 \pm 2.49$ & $2.67 \pm 0.94$ & $4.67 \pm 2.49$ & $5.33 \pm 2.49$ \\
            D0 & \textcolor{green}{\textbf{$97.33 \pm 0.94$}} & $91.33 \pm 0.94$ & \textcolor{green}{\textbf{$98.67 \pm 0.94$}} & $98.00 \pm 1.63$ \\
            D1 & \textcolor{green}{\textbf{$90.00 \pm 3.27$}} & $86.67 \pm 1.89$ & \textcolor{green}{\textbf{$94.00 \pm 1.63$}} & $88.67 \pm 3.40$ \\
            D2 & \textcolor{green}{\textbf{$87.33 \pm 2.49$}} & $86.67 \pm 0.94$ & $84.67 \pm 1.89$ & \textcolor{green}{\textbf{$85.33 \pm 4.11$}} \\
            \hline
            \textbf{MugCleanup} & & & & \\
            Source & $4.00 \pm 2.83$ & $4.00 \pm 1.63$ & $5.33 \pm 2.49$ & $2.00 \pm 0.00$ \\
            D0 & \textcolor{green}{\textbf{$95.33 \pm 1.63$}} & $94.67 \pm 2.49$ & \textcolor{green}{\textbf{$98.67 \pm 0.94$}} & $98.67 \pm 0.94$ \\
            D1 & $80.67 \pm 0.94$ & \textcolor{green}{\textbf{$82.67 \pm 2.49$}} & \textcolor{green}{\textbf{$96.67 \pm 0.94$}} & $95.33 \pm 0.94$ \\
            \bottomrule
        \end{tabular}
        \caption{Performance comparison of low-dim agents trained using Diffusion Policy and Behavior Cloning on datasets generated by DynaMimicGen (\textbf{D-MG}) versus those produced by the MimicGen (\textbf{MG}) baseline.}
        \label{tab:results_low_dim}
    \end{minipage}
\end{table}

Table \ref{tab:results_low_dim} summarize the results obtained using low-dimensional state observations for the \textit{Stack}, \textit{Square}, and \textit{MugCleanup} tasks. These results can be directly compared with the image-based agent results reported in Tables \ref{tab:results_image} to evaluate the impact of different observation modalities on policy performance.

We observe that agents trained with low-dimensional states consistently achieve higher success rates than their image-based counterparts.
This improvement is largely due to the reduced complexity of the input space: low-dimensional observations provide direct access to structured, task-relevant features, whereas image-based inputs are high-dimensional, computationally more demanding, and require additional representation learning.
Consequently, learning from low-dimensional inputs is more sample-efficient and allows policies to converge faster.

Importantly, these results demonstrate that datasets generated by DynaMimicGen (D-MG) retain critical task information across both low-dimensional and image-based representations, enabling effective policy learning regardless of the observation modality.
While the lower dimensionality of the state-based observations naturally boosts the performance of agents trained on MG datasets, D-MG still achieves superior success rates across the majority of tasks and distributions, especially with Diffusion Policy agents.
This confirms that D-MG’s adaptive data generation produces higher-quality, more informative demonstrations, which are robust to the choice of input representation.

Overall, these findings validate the versatility and generality of DynaMimicGen as a scalable data generation framework, capable of supporting effective policy training in both visual and state-based robotic control settings, while reducing the reliance on large numbers of human demonstrations.

\newpage

\section{Real Robot Experimental Details}
\label{app:real_robot_exp}

In this section, we discuss details on real environments, implementation and deployment of DynaMimicGen, and failure modes in real scenarios

\subsection{Real environments}

We evaluated DynaMimicGen (D-MG) on three distinct real-world robotic manipulation environments — Lift, Stack, and Cleanup — designed to progressively increase in complexity and contact dynamics.

The Lift task (Fig.~\ref{fig:real_envs}, top row) represents a basic manipulation scenario where the robot must approach and grasp a single cube, lifting it vertically from the table surface.
The task is considered successful if the cube is securely grasped and elevated to a designated height without slippage or collision.
This environment serves as a benchmark for assessing the system’s capability to generate and reproduce simple pick behaviors.

The Stack task (Fig.~\ref{fig:real_envs}, middle row) introduces greater spatial reasoning and precision requirements.
Here, the robot must approach, grasp, and lift a red cube before accurately placing it on top of a stationary purple cube.
The task is deemed successful if the red cube remains stably stacked on the purple one after placement.
This setup evaluates D-MG’s capacity to generalize to multi-step interactions involving contact stability and accurate alignment.

The Cleanup task (Fig.~\ref{fig:real_envs}, bottom row) represents the most challenging task, combining sequential, contact-rich, and long-horizon actions
The robot must first open a small box, then grasp a red cube, place it inside the box, and finally close the lid.
The task is successful if the cube is correctly deposited inside the box and the lid is fully closed.
This environment tests D-MG’s robustness in generating and executing temporally extended trajectories that require continuous adaptation to object dynamics and environmental constraints.

Collectively, these three environments provide a comprehensive evaluation framework, enabling us to assess D-MG’s scalability from simple pick-and-place motions to complex, multi-stage manipulation tasks under real-world physical conditions.

\subsection{DynaMimicGen implementation}

As described in Assumption 2, DynaMimicGen assumes the manipulation in each sub-task relative to a single object's coordinate frame $(o_{S_i} \in O)$.
Based on this, the models in the real environments are trained on observations relative only to the actual sub-task target.

For each sub-task $i$, the state is defined as $\mathcal{S}_i=\{X_{ee},R_{ee},X_{trg,i},R_{trg,i}\}$, where $i$ denotes the current sub-task, $X$ refers to the Cartesian positions in meters, and $R$ to the Cartesian orientations, expressed in quaternions.
This allows for robustness against occlusions and tag misdetections. 
The action space includes end-effector pose and gripper width, defined as $\mathcal{A}=\{X_{ee,trg}, R_{ee,trg},g\}$. Rotations are defined as axis-angle, as in \cite{mandlekar2023mimicgen}.

\subsection{Trajectory generation}

To create a dataset with generated trajectories in the real environments, we run multiple trials, recording data from the successful ones.

A single initial human demonstration is collected for each environment, in a static configuration.
For each environment, the full demonstration is composed of a sequence of smaller demonstrations that define the sub-tasks (e.g., approach, stack, open box, etc.). 
In this way, demonstration record and sub-task annotations happen simultaneously.

Each sub-task is mapped with the proper sub-demonstration, and each task is composed of a defined sequence of sub-tasks.
A DMP is associated with each sub-task and trained with the corresponding sub-demonstration.

It is then possible to run the trajectories generation process by setting up the correct environment. 
At run time, DynaMimicGen monitors the state of the task, based on the ongoing sub-task, defines the target pose, and adapts the robot end-effector actions and gripper opening to the current, dynamic environment's state.
While the robot is moving, the user modifies the objects' positions to make the environment dynamic.

State-Action pairs and other relevant information are stored in a format compatible with RoboMimic \cite{robomimic2021} for training deep learning algorithms.

\subsection{Training agents with more generated trajectories}
\label{ssec:bc_more_data}

To assess the statement in \ref{subsec:real_robot_exps} that more generated trajectories would result in better performances of the trained BC agent, we generated 200 trajectories for the Lift case and trained BC with this dataset.
Experimental results confirm this assumption, allowing the trained model to reach 96.0\% of success rate over 50 trials, showing a relative improvement of more than 11\% compared to the model trained with 50 generated demos.
The two failures happened in one case because the square was moved outside the training workspace, and in the other because the vision system failed to properly localize the square due to poor illumination when the robot moved above it.

This result also reflects the advantage of training an effective BC agent instead of using vanilla DMPs as a motion policy. Indeed, the BC agent performs well even if the target object is moved when the robot is close to it, ready for grasping. In this case, DMPs do not allow for replanning, as they are fixed length, and they can't recover once they finish. On the contrary, the BC agent can re-plan based on the updated state and adapt the task's length until solved.
This behavior is visible in the video attached.

\subsection{Failure modes in the real scenario}
\label{ssec:failure_modes_real}

Here we present the most common failure modes noted during testing. Failures happen both in trajectory generation and running BC agent, in different modalities, for different causes.

\subsubsection{Trajectories generation failure modes}

During trajectory generations, we observed failures in solving the tasks, happening mostly due to i) tag occlusions or misdetections, ii) imperfect camera calibration, and iii) too dynamic environments, in particular in the final phase of sub-tasks.

Concerning i), the choice of using ARUCO markers was made to make object localization straightforward. Despite this, it also introduced possible failures, and the choice of a different detection method (e.g., FoundationPose \cite{wen2024foundationpose}) can possibly increase robustness. 
Indeed, markers need to be always completely visible and properly lit. If any occlusion, even partial, occurs, the target is no longer detected, and the pose cannot be estimated. Moreover, in certain cases, the marker falls in the robot's shade, reducing lighting and causing possible misdetections.

Vision systems require proper camera-to-base calibration. We calibrated intrinsic camera parameters to correct distortions and performed eye-to-base calibration using the EasyHandEye package.
Despite this, small uncertainties are still present, if compared with the simulated environments that allow full and precise state knowledge in every situation.
In particular, we observed failures when the target objects were placed close to the borders of the camera's field of view, possibly due to minor distortions or small hand-eye calibration errors.

Finally, DMPs excel in reproducing trajectories' shape and have been proven to adapt to dynamic environments, allowing online adaptations of the trajectory. Despite this, they also have a fixed length, meaning that if the task is not completed in their allowed steps, they fail to solve it.
We observed that, during data collection, if the object is moved close to the end of the sub-task, DMPs fail to recover from the updated state and cannot solve the task properly. 
Similarly, if changes in the environment are too fast, DMPs cannot always properly track the target position.
This issue can be partially overcome by a different tuning of DMPs parameters, making the robot's behavior more reactive in tracking the target.
Despite this, a too reactive robot results in jerky motions and possibly unfeasible motions due to intrinsic safety constraints in collaborative robots.

\subsubsection{BC failure modes}

Failures in BC usage are due mostly to i) few data for training, ii) target poses outside training space, and iii) tag detection and calibration issues.

The main reason BC has lower success rates than generation success rates is that BCs were trained on few data. We collected 50 generated trajectories for each environment, which is way less data than in the simulated environments experiments. 
Therefore, despite BC showing promising results, its training cannot generalize well with the small amount of data provided, and failures occurred.
This is also confirmed by the results obtained in \ref{ssec:bc_more_data}.

Moreover, during testing of BC, happened that the target object fell at the border or even out of the training space, reducing the capabilities of BC to properly handle states outside the training distribution.

Finally, we observed the same issues related to the vision system and tag detection, as described in the previous subsection.

\subsection{Sub-task success rates}

Here we present the success rates observed for each of the sub-tasks, observed when evaluating BC.

Table \ref{tab:real_exps_st} shows success rates for each sub-task.
The Lift environment ends when the cube is picked, and its success rate is the same as from \ref{tab:real_exps}.

For the Stack environment, we observe a clear drop in performance from the pick to the stack sub-tasks.
Most of the failures were due to the improper placement of the cube above the other, leading the lifted cube to fall down, rolling onto the bottom one.
Misalignments are due to i) vision system uncertainties and occlusions, ii) the gripper opened a bit earlier than proper positioning, and iii) the target positions close to the training space boundaries. 

In the Cleanup environment, we observed a high success rate in opening the box, with a few failures due to the box getting stuck in the gripper.
The high success rate, even higher than cube pick, reflects the fact that the box was moved in a smaller workspace compared to that of the Lift environment. 
The success rate in picking the object reduces, and failures were observed at the beginning of the task, with the robot showing a "lost" behavior. This is possibly because after box opening the robot and the cube are in a state outside the training distribution. A similar behavior was also observed for the place and close box sub-tasks.

Success rates can be improved with more training data, as demonstrated in \ref{ssec:bc_more_data}.

\begin{table}[h]
    \centering
        \centering
        \begin{tabular}{lcccc}
         & open box & pick & stack/place & close box \\
        \hline
        Lift  & N/A & 86.0\%   & N/A & N/A\\
        Stack  & N/A & 84.0\%  & 32.0\% &  N/A \\
        Cleanup & 96.0\% & 68.0\%   & 42.0\% & 26.0\%\\
        \end{tabular}
        \caption{\textbf{Sub-task success rates} Success rates of sub-task measured while evaluating BC in \ref{subsec:real_robot_exps}. }
        \label{tab:real_exps_st}
\end{table}

\newpage

\section[\appendixname~\thesection]{Limitations and Failure Modes}
\label{app:limitations}

In this section, we discuss the limitations of D-MG that can motivate and inform future works.
\begin{enumerate}
    \item \textbf{Known sequence of object-centric substaks.} 
    As done in \cite{mandlekar2023mimicgen}, D-MG assumes knowledge of the object-centric subtasks in a task (which object is involved at each subtask) and also assumes that this sequence of subtasks does not change (Assumption 2, Sec. \ref{sec:problem}).
    \item \textbf{Known object poses at each time-step of each subtask during data generation.} During data generation, at eahcc time-step of each object-centric subtask, D-MG requires an object pose estimate of the reference object for that subtask (Assumption 3, Sec \ref{sec:problem}).
    This assumption is essential to continuously adapts the robot's actions based on real-time feedback from the environment, allowing for dynamic re-planning when unexpected events occur (for more detail see Appendix \ref{eq: dmp_equation}).
    \item \textbf{One reference object per subtask.}
    RoboDatGen assumes each task is composed of a sequence of subtasks that are each relative to exactly one object (Assumption 2, Sec. \ref{sec:problem}).
    Being able to support subtasks where the motion depends on more than one object (for example, placing an object relative to two objects, or on a cluttered shelf) is left for future work.
    \item \textbf{No Obstacle Integration in Trajectory Generation.}
    While DynaMimicGen (D-MG) effectively enables large-scale dataset generation in both static and dynamic environments, it currently does not account for the presence of obstacles. Future work will explore data generation methods for robotic manipulation tasks in environments with varying numbers and configurations of obstacles. One promising direction involves incorporating coupling terms into the DMP formulation to manage contacts and obstacle avoidance, as done in \cite{5152423}.
    \item \textbf{No support for multi-arm tasks.}
    D-MG only works for single arm tasks and it is not intended to be used to generate datasets for multi-manual manipulation.
    \item \textbf{Limited sim-to-real experiments.}
    Investigating the sim-to-real transfer of policies trained on D-MG generated datasets remains an important direction for future work.
    Fortunately, our framework is highly compatible with existing sim-to-real methodologies \cite{peng2018sim, kaspar2020sim2real, zhao2020sim}, many of which have shown considerable success in transferring imitation-learned policies from simulation to the real world.
    Recent advances in sim-to-real learning — such as those described in \cite{peng2018sim, kaspar2020sim2real, zhao2020sim} — demonstrate the feasibility of bridging the sim-to-real gap, particularly when supported by robust data generation strategies and transfer techniques.
    These include domain randomization \cite{tobin2017domain}, a widely used approach that perturbs various elements of the simulated environment (\textit{e.g.}, lighting, textures, dynamics) to improve generalization to real-world settings.
    D-MG is fully compatible with such techniques, making it a promising candidate for realistic policy deployment on physical robots.\\
    Importantly, our current focus has been on building a scalable, flexible, and general-purpose data generation pipeline in simulation, where data can be generated in large quantities without the constraints typically associated with real-world data collection — such as slow resets, limited availability of hardware, and the need for human supervision.
    In future work, we aim to investigate how D-MG-generated data can be integrated with sim-to-real adaptation pipelines to train and deploy robust policies on real robots, paving the way for practical deployment in real-world manipulation tasks.
    \item \textbf{Failure modes in dynamic environment.}
    While DynaMimicGen remains effective in dynamic settings, Table \ref{tab:dyn_gen} reveals a notable reduction in Data Generation Success Rates (DGRs) compared to the higher rates achieved under static conditions reported in Table \ref{tab:DMGvsMG}. This decline stems from inherent limitations in how Dynamic Movement Primitives (DMPs) are applied during demonstration generation.
    Specifically, each DMP-generated trajectory is constrained to match the duration of the original demonstration, which assumes a fixed number of timesteps. Consequently, when dynamic perturbations—such as sudden changes in object positions—occur late in the trajectory execution (e.g., immediately before a grasping action), the DMP controller lacks sufficient time to adequately adapt the remaining trajectory to the new conditions.
    This temporal inflexibility means that the robot may miss or improperly execute critical manipulation actions, such as object grasping or placement, leading to task failure. As a result, the overall DGR is negatively affected in dynamic scenarios. These observations highlight the importance of developing more adaptive or time-flexible trajectory generation mechanisms that can accommodate unexpected changes in task conditions, particularly when those changes occur during time-sensitive sub-tasks.
    \item \textbf{Failure modes in real-world experiments.} In the real-world experiments, failures happen both at trajectory generation and when running BC trained models. 

    In particular, real-world scenarios introduce object pose uncertainties compared with the simulations. Moreover, object pose may not always be visible due to poor lighting or occlusions. Clearly, with an accurate camera calibration, proper lighting, and good positioning of the vision system (possibly with more cameras placed at different positions), uncertainties in pose estimation are reduced.
    
    Specific to trajectory generations, failures also happen because of a too dynamic environment, particularly in the final phase of sub-tasks. If the object is moved close to the end of the DMP trajectory, the DMP cannot adapt properly to the new state.
    
    When running BC models, specific failures are due mostly to i) few data for training, ii) target poses outside the training space.
    More data would improve the robustness of BC models, while training on data collected for the full robot's workspace allows it to reduce the second failure mode.

    More details on the failures for the real use-cases are in \ref{ssec:failure_modes_real}.
\end{enumerate}

\newpage

\section*{Acknowledgments}
The research has been funded by the EU project FLUENTLY. Grant agreement no 101058680.

\bibliographystyle{unsrt}  
\bibliography{references}

\end{document}